\documentclass{article}

\usepackage{arxiv}

\usepackage[utf8]{inputenc} %
\usepackage[T1]{fontenc}    %
\usepackage{hyperref}       %
\usepackage{url}            %
\usepackage{booktabs}       %
\usepackage{amsfonts}       %
\usepackage{nicefrac}       %
\usepackage{microtype}      %
\usepackage{lipsum}		%
\usepackage{graphicx}
\usepackage{natbib}
\usepackage{doi}
\usepackage{amsmath}

\usepackage{caption}
\usepackage{tikz}
\usepackage[font=scriptsize]{subcaption}
\usepackage{xcolor}
\usepackage{xspace}

\title{Learning Deep Morphological Networks with Neural Architecture Search}

\author{ Yufei Hu\\
U2IS, ENSTA Paris, 
Institut Polytechnique de Paris\\
	\texttt{yufei.hu.2021@ensta-paris.fr} \\
	\And
	Nacim  Belkhir\\
	Safrantech, Safran Group\\
	\texttt{nacim.belkhir@safrangroup.com} \\
		\And
	Jesus Angulo\\
	CMM, Mines ParisTech, PSL Research University\\
	\texttt{jesus.angulo@mines-paristech.fr} \\
		\And
	Angela  Yao\\
	School of Computing National University of Singapore\\
	\texttt{ayao@comp.nus.edu.sg} \\

\And
Gianni Franchi\\
U2IS, ENSTA Paris, 
Institut Polytechnique de Paris\\
	\texttt{gianni.franchi@ensta-paris.fr} \\
}

\renewcommand{\shorttitle}{Learning Deep Morphological Networks with Neural Architecture Search}

\newcommand{\Gianni}{\textcolor{black}}

\newcommand{\Correction}{\textcolor{black}}
\newcommand{\Newcorrection}{\textcolor{black}}

\makeatletter
\DeclareRobustCommand\onedot{\futurelet\@let@token\@onedot}
\def\@onedot{\ifx\@let@token.\else.\null\fi\xspace}

\def\eg{\emph{e.g}\onedot} 
\def\ie{\emph{i.e}\onedot}

\makeatother

\begin{document}
\maketitle

\begin{abstract}
Deep Neural Networks (DNNs) are generated by sequentially performing linear and non-linear processes. The combination of linear and non-linear procedures is critical for generating a sufficiently deep feature space. Most non-linear operators are derivations of activation functions or pooling functions. Mathematical morphology is a branch of mathematics that provides non-linear operators for various image processing problems. This paper investigates the utility of integrating these operations into an end-to-end deep learning framework. DNNs are designed to acquire a realistic representation for a particular job. Morphological operators give topological descriptors that convey salient information about the shapes of objects depicted in images.
We propose a method based on meta-learning to incorporate morphological operators into DNNs.
The learned architecture demonstrates how our novel morphological operations significantly increase DNN performance on various tasks, including picture classification, edge detection, \Newcorrection{and semantic segmentation. Our codes are available at \url{https://nao-morpho.github.io/}}.
\end{abstract}


\section{Introduction}
Over the last decade, deep learning has made several breakthroughs and demonstrated successful applications in various fields (\eg  computer vision \Correction{\cite{alexnet,vggnet, resnet}}, object detection \cite{yolo}, and NLP \Correction{\cite{ GPT}}). This success is mainly attributable to the fact that the feature engineering process is automated, whereby features are learned in an end-to-end process from data rather than designed manually. The need for improved architecture has swiftly followed the advent of deep learning, with experts now placing a premium on architecture engineering \emph{in lieu of} feature engineering.

Architecture engineering is concerned with determining the most appropriate operations for a network, their hyper-parameters (\eg the number of neurons for fully connected layers or the number of filters or the kernel size for convolutional layers), and the connectivity of all operations. Generally, practitioners propose novel operations to validate various architectures and tasks to improve performance on specific tasks. As a result, developing a novel operation remains a time-consuming and costly process that necessitates a manual search for the optimal configuration. It is, therefore, prone to failure when practitioners lack computational resources. An alternative approach focuses on automatically finding the network architecture design using Neural Architecture Search (NAS) methods \cite{liu2019darts,luo2019neural,liu2018progressive,pham2018efficient,8681706} in lieu of manual design. Given a set of data and a performance metric for a learning task, an NAS algorithm attempts to find the optimal architecture for a search strategy. It can be viewed as an optimization problem in the space of an architecture network defined by a collection of operations and their possible combinations. Recently, it was demonstrated empirically on several applications that architectures discovered using NAS outperform those discovered manually.

We are observing steadily increasing interest in mathematical morphology among the deep learning community. Indeed, the intrinsic features of mathematical morphology operators that enable them to extract information from topological structures make them excellent candidates.
Morphological operators have been shown to capture image edges \cite{rivest1993morphological}, granulometry~\cite{serra1988image,thibault2013advanced}, and distances to object borders~\cite{franchi2014comparative}. 
The literature primarily employs two methodologies to evaluate the utility of morphological operators. Some analyses ~\cite{cavallaro2017automatic,velasco2013classification, franchi2016morphological} directly extract descriptors from unlearned morphological layers, while others ~\cite{FRANCHI2020107246,Valle_2020,mondal2020image} propose learning the structural element of the morphological operators. Despite their superior performance in various applications, these methodologies are prone to failure if the deep network architecture is misdesigned, which may dissuade researchers from pursuing this research path.

\begin{figure}[htbp]
  \centering
  \includegraphics[width=0.6\textwidth]{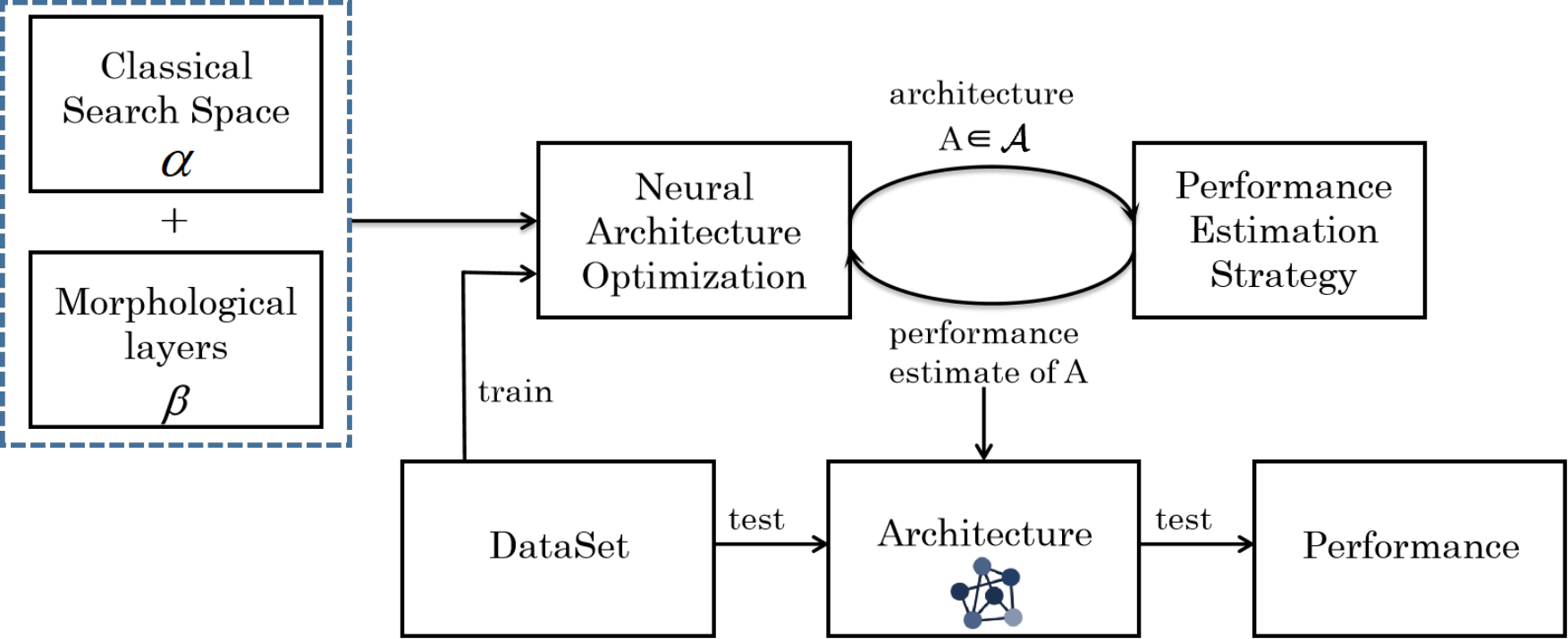} 
  \caption{\Correction{The overall methodology based on neural architecture search to assess the usefulness of morphological layers.}} 
  \label{overallpaper}
\end{figure}

\Correction{This paper proposes a novel methodology based on NAS to assess the usefulness of newly developed architecture layers and, in particular, morphological layers (see Figure \ref{overallpaper})}.
The following empirically investigates morphological layers applied to deep networks on CIFAR10/CIFAR100 \cite{Krizhevsky09learningmultiple} an edge detection task using BSD500 \cite{MartinFTM01}, and a semantic segmentation task on Cityscapes \cite{cityscapes} to determine the optimal design for morphological layers. We perform a comparison between our results and the best architecture discovered using conventional convolutional layers.

Our contributions can be summarized as follows:
\begin{itemize}
\item First, we propose novel procedures based on sub-pixel convolutions and mathematical morphology to construct pseudo morphological operations using standard convolution layers.  %
\item We integrate these procedures into deep networks using morphological layers and NAS algorithms. We demonstrate empirically that our %
architecture tailored to morphological layers can outperform conventional convolutional layers.
\item We outline some current issues in NAS and introduce the problem of choosing the backbone, \ie the higher-level architecture design on which the search will be performed. We offer novel network space descriptions suitable for the edge identification task.
\item We are the first to examine architectural search mixed with morphological procedures for edge detection \Newcorrection{and semantic segmentation}. Our new specialized architecture achieves state-of-the-art performance in edge detection.

\end{itemize}

\section{Related work}

\subsection{Mathematical Morphology}

Mathematical morphology has been extensively used to denoise raw images \cite{Jserra,bouchet2016fuzzy} as well as characterize and analyze microscopic images \Correction{\cite{franchi2018enhanced}} and remote sensing images~\cite{franchi2016morphological,franchi2014comparative, cavallaro2016spectral,cavallaro2017automatic,velasco2013classification}. Additionally, these operators have been employed to generate medical images  \cite{dufour2013morphology,zhang2012spatial}. However, all of these morphological paper operators were used as filters to derive feature descriptors from the classifier's input data. Here, we integrate them into a DNN.

\subsection{Morphological Neural Network}

Masci et al.\cite{masci2013learning} pioneered the use of morphological operators in neural networks by researching pseudo harmonic morphological operators in conjunction with DNNs. The first morphological Perceptrons were investigated in  \Correction{\cite{saeedan2018detail,mondal2019dense,Valle_2020,charisopoulos2017morphological}}. Some investigations attempted to integrate morphological layers and DNN architecture
\cite{mellouli2017morph,mondal2020image,FRANCHI2020107246,nogueira2019introduction}. The issue with these works is that the effectiveness of morphological layers are arguably architecture-dependent, and hence, our work proposes linking the search for architecture using morphological layers.
We hereby disregard the work of Blusseau et al. \cite{Blusseau2020}, who used auto-encoder networks to approximate morphological operators.

\subsection{Neural Architecture Search} 
Experts in the machine learning field typically create DNNs by hand and select hyper-parameters through trial-and-error, making the process tiring and tedious as well as prone to errors. A different perspective sees model design as a decision-making process that can be improved wherein we can automatically find the best combination of algorithms to maximize the performance of a task. Amidst the growing interest in deep learning, AutoML \cite{survey_on_automl} and NAS~\Gianni{\cite{liu2018progressive}} have emerged, whereby the entire DL pipeline can be automated, aiming to reduce the overall development cost and approach expert performance.
There have been several efforts \Correction{\cite{ pham2018efficient, luo2019neural}}, as early as the 1990s \Correction{\cite{kitano1990designing}}, to formulate NAS as an optimization in the space of network architectures, solved using either reinforcement learning algorithms \cite{pham2018efficient}, gradient-based optimization~\cite{liu2019darts}, sequential model-based optimization \Correction{  \cite{luo2019neural, camero2020bayesian}, or evolutionary algorithms \cite{liu2021survey,loshchilov2016cma}. 
While several search strategies have been investigated, gradient-based and sequential-based methods appear to reach state-of-the-art performance with lower computational cost \cite{comprehensive_nas}. Despite good performance, agent-based methods \cite{zoph2018learning} and evolutionary algorithms remain expensive as they require time and effort to reach a good candidate solution \cite{DBLP:journals/corr/abs-2101-07415}. Moreover, as explained in \cite{eiben2011parameter}, evolutionary algorithms are highly sensitive to the hyperparameters.
Here, we use NAO \cite{luo2019neural}, which is a good compromise.}

\section{Morphological Architecture Search (MAS)}~

We first outline some preliminaries on the convolution operation and its relationship to mathematical morphology in Sec.~\ref{sec:prelim}. We then outline our proposed pseudo morphological dilation operation (Sec.~\ref{sec:layerdilation}) and its variants (Sec.~\ref{sec:layerpresentation}). Next, we explain the architecture search algorithm that we use to integrate the operators into a neural network (Sec.~\ref{sec:NAS}). Finally, we describe the proposed architecture backbone. 

\subsection{Preliminaries}~\label{sec:prelim} 

Consider a discrete RGB image $f$, where $f[i,j,0],f[i,j,1],f[i,j,2]$ denote the red, green, and blue values at position $(i,j)$, respectively. We further denote $g$ as the feature map resulting from a DNN's convolution layer of $f$ with the filter $\omega$ without bias. The feature map $g$ can be expressed as: 

\begin{eqnarray}
 g[n_1,n_2,c]= \sum _{k=0}^{2}  \sum _{(i,j) \in \mathcal{N}} f[n_1 +i,n_2+j,k]\omega[i,j,k] + b[c] \mbox{ , with } c \in [1,C_{\mbox{out}}] 
\end{eqnarray}

\noindent where $\mathcal{N}$ is a square kernel that defines the spatial size of the convolution kernel, $c$ is the index of the channel, $C_{\mbox{out}}$ is the number of channels output by the layer, and $b$ is the bias, which is equal to zero if we consider unbiased convolution. %

\begin{figure}[!t]
\begin{center}
        \centering
        \subfloat[grayscale]{\begin{minipage}[t]{0.15\linewidth}
        \centering
        \includegraphics[width=0.9\linewidth]{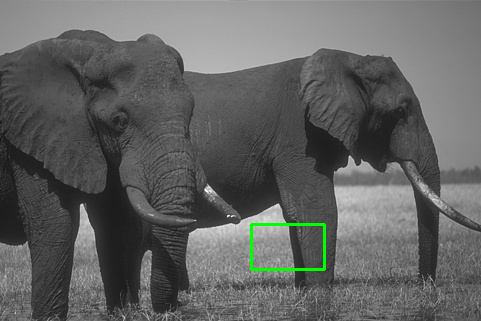}\label{original image}
        \end{minipage}}
        \subfloat[ dilation]{\begin{minipage}[t]{0.15\linewidth}
        \centering
        \includegraphics[width=0.9\linewidth]{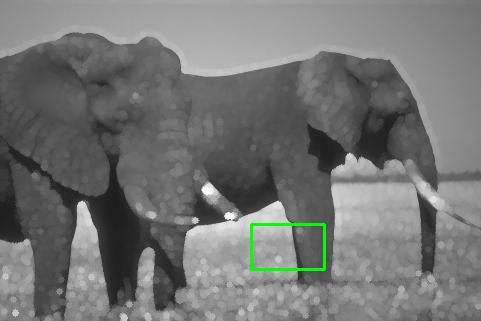}\label{dilation}
        \end{minipage}}
        \subfloat[ erosion]{\begin{minipage}[t]{0.15\linewidth}
        \centering
        \includegraphics[width=0.9\linewidth]{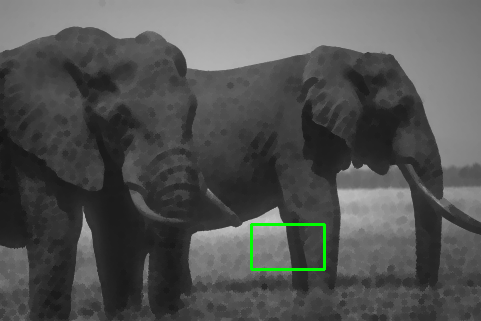}\label{erosion}
        \end{minipage}}
        \subfloat[gradient]{\begin{minipage}[t]{0.15\linewidth}
        \centering
        \includegraphics[width=0.9\linewidth]{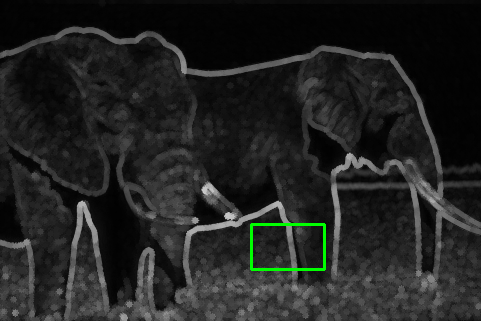}\label{gradient}
        \end{minipage}}
        \subfloat[opening]{\begin{minipage}[t]{0.15\linewidth}
        \centering
        \includegraphics[width=0.9\linewidth]{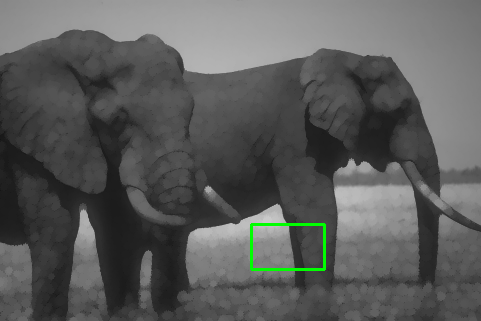}\label{opening}
        \end{minipage}}
        \subfloat[closing]{\begin{minipage}[t]{0.15\linewidth}
        \centering
        \includegraphics[width=0.9\linewidth]{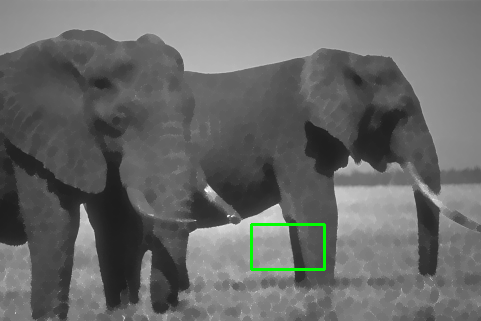}\label{closing}
        \end{minipage}}
        
        \subfloat[grayscale]{\begin{minipage}[t]{0.15\linewidth}
        \centering
        \includegraphics[width=0.9\linewidth]{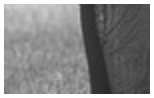}\label{ scale original image}
        \end{minipage}}
        \subfloat[ dilation]{\begin{minipage}[t]{0.15\linewidth}
        \centering
        \includegraphics[width=0.9\linewidth]{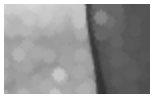}\label{scale dilation}
        \end{minipage}}
        \subfloat[ erosion]{\begin{minipage}[t]{0.15\linewidth}
        \centering
        \includegraphics[width=0.9\linewidth]{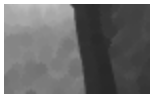}\label{scale erosion}
        \end{minipage}}
        \subfloat[gradient]{\begin{minipage}[t]{0.15\linewidth}
        \centering
        \includegraphics[width=0.9\linewidth]{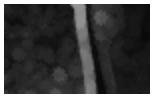}\label{scale gradient}
        \end{minipage}}
        \subfloat[opening]{\begin{minipage}[t]{0.15\linewidth}
        \centering
        \includegraphics[width=0.9\linewidth]{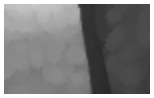}\label{scale opening}
        \end{minipage}}
        \subfloat[closing]{\begin{minipage}[t]{0.15\linewidth}
        \centering
        \includegraphics[width=0.9\linewidth]{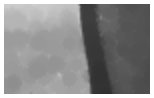}\label{scale closing}
        \end{minipage}}
        
        \end{center}
         \caption{Dilation and erosion transformation where the structuring element is disk (3) applied to a sample image from the BSD500 dataset~\cite{amfm_pami2011}. %
         }
     \label{DE}
         \end{figure}

By analogy, mathematical morphology~\cite{Jserra} operators are non-linear image operators based on the spatial structure of the image. Initially, these operators were proposed for binary images, but have now been extended to grayscale images. 
Let $f$ be a grayscale image representing a function, with the intensity at position $x$ denoted as $f(x)$. The two basic operations in morphology are performed at the gray level, such that we define the erosion and dilation operations on their discrete version respectively as:  $\varepsilon_{b}(f)[n_1,n_2]= \min_{(i,j) \in \mathcal{N}} f[n_1 +i,n_2+j] - b[i,j]$ and 
$\varepsilon_{b}(f)[n_1,n_2]= \max_{(i,j) \in \mathcal{N}} f[n_1 -i,n_2-j] + b[i,j]$,
where $b=\left[b_{-n} , \ldots, b_{n}\right]$  is a structuring element (SE). %
In Figure \ref{DE}, we observe that the dilation increases the bright areas based on the shape of the SE,
 leading to a brighter image.
Erosion is the morphological dual to dilation and decreases the bright areas.
 We can see a direct link between the dilation/erosion and convolution, whereby the dilation is a convolution in the max + algebra, as explained in \cite{angulo2017convolution}. 

These operations are increasing, hence: $f<g \Rightarrow \varepsilon_{b}(f)(x)<\varepsilon_{b}(f)(x)$ or$f<g \Rightarrow \delta_{b}(f)(x)<\delta_{b}(f)(x)$. In addition, the erosion is anti-extensive, while the dilation is extensive, hence:$f>=\varepsilon_{b}(f)(x)$ and $\delta_{b}(f)(x)>=f$.

By combining these two basic operations, we can build new ones, such as the opening and closing. The opening of image $f$ by structuring element $b$ is given by applying an erosion on $f$ with the structuring element $b$ and then applying a dilation to the previous results with the same SE. The closing operation is the morphological dual to the opening. 
\begin{eqnarray}\label{OpeningClosing}
  \gamma_{b}(f) =  \delta_{b}\left( \varepsilon_{b}(f) \right)\mbox{ , and   } 
  \varphi_{b}(f)  =  \varepsilon_{b}\left( \delta_{b}(f) \right).
\end{eqnarray}

By combining these two basic operations, we can also build  the internal gradient and the external gradient, denoted respectively as $G_i$ and $G_e$ and defined as:
$
    G_e(f) = \delta_{b}(f)-f\mbox{ , and  }
    G_i(f) = f-\varepsilon_{b}(f)
$.

In addition, we can also build the morphological gradient, which is the difference between the dilation and the erosion with the same SE applied to the same image. Hence, it is defined as: 
$
    G_b(f) = \delta_{b}(f)-\varepsilon_{b}(f)
$.
The structuring element impacts the morphological operations through both the geometry of its support and its weights. 
Hence, by combining the morphological operators, we can build new operators. The question arises as to how to combine these operators to achieve the best performance for a given task. These operators have long been combined based on expert knowledge, but in this work, we propose combining them using an architecture search algorithm.

\subsection{Our Pseudo Morphological Dilation}\label{sec:layerdilation}

\begin{figure}[htbp]
\centering
\begin{minipage}{.55\columnwidth}
    \subfloat[Overview of the projection convolution, where r equals 2]{
      \label{pseudo_shuffle_a} 
      \includegraphics[width=.55\textwidth]{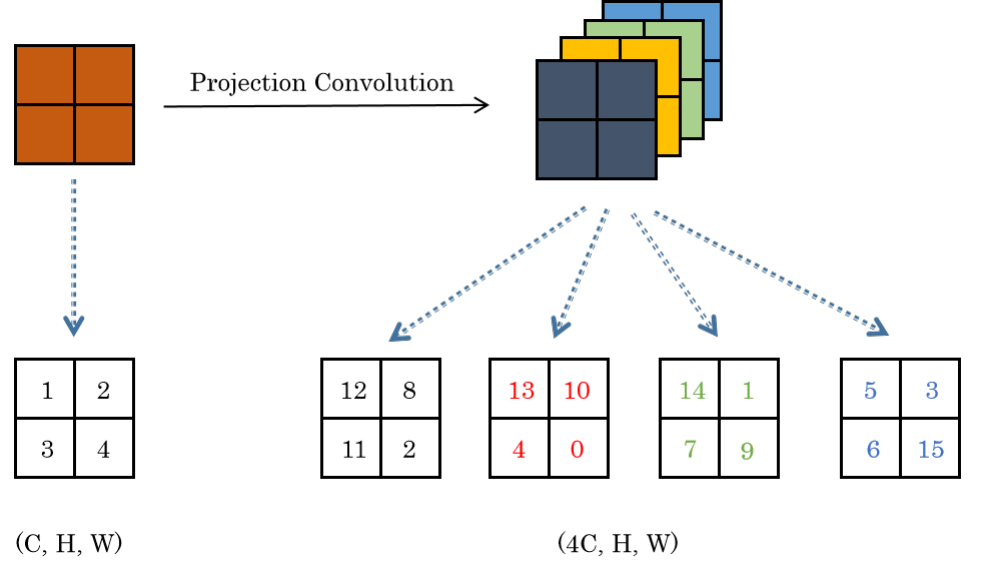}}\\
    \subfloat[Overview of the pixel shuffle]{
      \label{pseudo_shuffle_b} 
      \includegraphics[width=.55\textwidth]{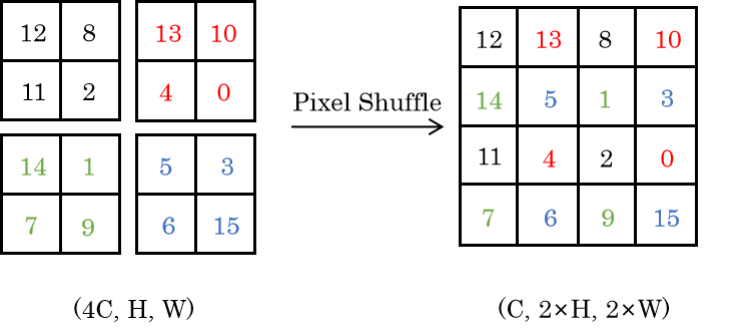}}
\end{minipage}
\caption{A simple example to explain the projection convolution and pixel shuffle step. We start with a feature map composed of 4 pixels of value (1,2,3,4) and then apply the projection convolution, with four feature maps of values (12,8,11,2),(13,10,4,0),(14,1,7,9) and (5,3,6,15). Then, by applying the pixel shuffle, we obtain just one feature map.}\label{pseudo}
\end{figure}

We propose the \textbf{pseudo morphological dilation} operation, which is composed of the following steps:

\begin{enumerate}
    \item We apply a traditional convolution operation to transform the feature map \Newcorrection{from} $(C, H, W)$ into $(C \times r \times r, H, W)$, where $r$ represents the size of the convolution kernel, and $H$ and $W$ respectively represent the height and width of the image.
 Hence, we increase the number of feature map channels. Each feature map represents a neighborhood map that will be used in the step. For example, in Figure \ref{pseudo_shuffle_a}, we have four feature maps. We denote this step the \textbf{projection convolution step.} Let us denote $f$ as the input image and $\omega$ as the convolution kernel of spatial size $1 \times 1$, and $b$ as the bias of the projection convolution. Let us consider that the convolutional layer has $C_{\mbox{in}}$ input channels and $C_{\mbox{out}}$ output channels. Hence, the resulting feature map at pixel $n_1,n_2$ is equal to: 
\begin{eqnarray}\label{proj_step}
 g[n_1,n_2,c]=\sum _{k=0}^{C_{\mbox{in}} } f[n_1,n_2,k]\omega[0,0,k] + b[c] \mbox{ , with } k_2 \in [1,C_{\mbox{out}}] 
\end{eqnarray}
 
 \item We then apply the  pixel shuffle transformation to the feature maps representing the coordinates, as illustrated in Figure \ref{pseudo_shuffle_b}. The pixel shuffle transformation, also known as the sub-pixel convolution, was introduced in \cite{shi2016realtime} for the super-resolution task. The pixel shuffle transformation reorganizes the low-resolution image channels to obtain a bigger image with fewer channels. Specifically, it increases the spatial size of the feature map by reducing the number of channels. Hence, it rearranges the input tensor elements, expressed as $(C \times r \times r, H, W)$, to form a scaled $(C, H \times r , W \times r )$. This operation is interesting since it is stable, compatible with deep learning back propagation, and does not add artifacts. We denote this step the \textbf{pixel shuffle step}.
More formally, each channel $c$ of the previous step represents a neighborhood, as illustrated in Figure \ref{pseudo}. Let us decompose equation \ref{proj_step} into two terms such that for all $c$ we have $ g[n_1,n_2,c]=\tilde{f}[n_1,n_2,c] + b[c]$, where $\tilde{f}$ is the results of the cross correlation of the convolution layer. Note that $C_{\mbox{out}} = r^2 C_{\mbox{in}}$. The output of the pixel shuffle is:
\begin{eqnarray}
 h[n_1,n_2,c]= g[ \lfloor n_1/ r \rfloor ,\lfloor n_2/ r \rfloor ,c']
\end{eqnarray}
where $\lfloor n_1/ r \rfloor$ is the floor fraction, which takes as input a real number $n_1/ r$ and outputs the greatest integer value. The channel $c'$ is equal to $c'=c+n_1\mbox{.mod}(r)+n_2\mbox{.mod}(r)+r\mathbf{1}_{n_2\mbox{.mod}(r)>0}$, where :$ n_1{\mbox{.mod}(r)}=n_1-(\lfloor n_1/r\rfloor \times r)$.
 \item On the output of the pixel shuffle step, we apply a max-pooling of
stride $r$.  We denote this step the \textbf{max-pooling step.}
We apply the max-pooling into $h$ with a kernel of spatial size $r \times r $ and a stride $r$; let us denote $h'$ as the result. Then, $h'$ is equal to: 
\small
\begin{eqnarray}
 h'[ n_1  ,n_2 ,c]= \max_{(i,j) \in \mathcal{N}} \tilde{f}[\lfloor (n_1 +i)/ r\rfloor ,\lfloor(n_2 +j)/ r\rfloor,c'[i,j]] + b[c'[i,j]] 
\end{eqnarray}
\normalsize
where $c'[i,j] = c+(n_1+i)\mbox{.mod}(r)+(n_2+j)\mbox{.mod}(r)+r\mathbf{1}_{(n_2+j)\mbox{.mod}(r)>0}$. Thus, we obtain a dilation on $\tilde{f}$, where the bias $b$ is the structuring element, with the connectivity depending on the channel.  

\end{enumerate}

Classical morphological operators can be hard to train due to the non-linearity, as pointed out in~\cite{FRANCHI2020107246}, where the authors proposed clipping the gradients of these layers and applying specific learning to them. Hence, integrating them into a NAS framework can be challenging as the NAS framework will try to learn the best architecture with the set of operations. Thus, we cannot build a hand-designed architecture that will stabilize the new layers, and the new layers must be stable for any architecture. Therefore, we propose a stable version, as introduced in this section.

\begin{figure}[htbp]
  \centering
  \includegraphics[width=.55\textwidth]{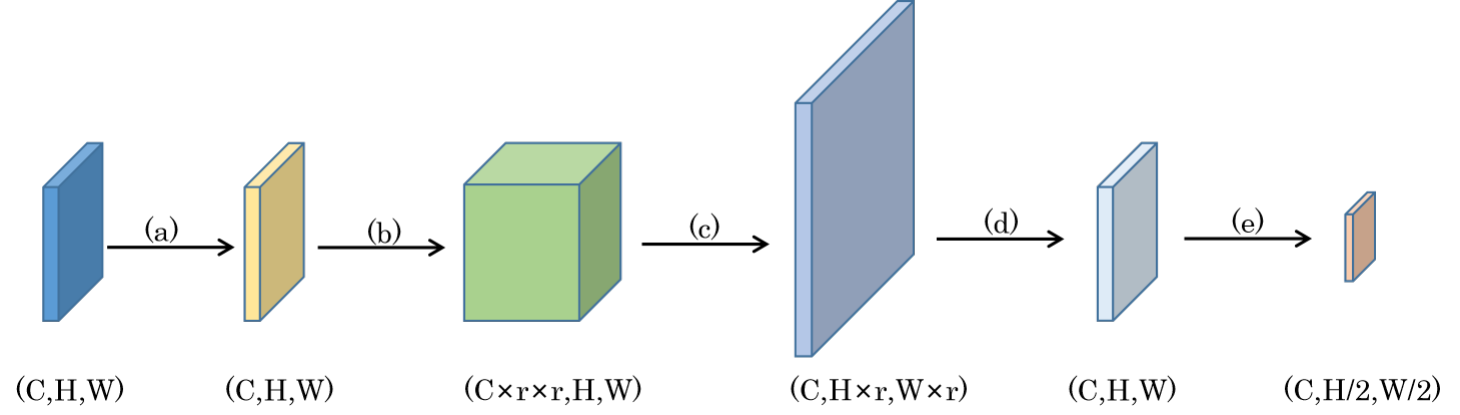} 
  \caption{An overview of \textbf{pseudo morphological dilation, pool}, and \textbf{erosion}. The transformation is composed of 4 mandatory steps (a),(b),(c),(d) and one optional step (e).
  (a) Batch normalisation to transform the input feature maps. (b) Traditional convolution to transform the feature map from (C, H W) into (C×r×r, H, W). (c) A
sub-pixel convolution that makes the length and weight of the feature map expand by r times. (d) Max-pooling or min-poling if we perform a dilation or an erosion, respectively, that makes the output feature map dimension
consistent with the input feature map.
(e) An optional max-pooling operation with a stride of 2.}
  \label{pseudo_mor_shuff}
\end{figure}

The full layer is represented in Figure \ref{pseudo_mor_shuff}.
We simulate the structuring element's shape by working with the parameter $r$ of the max-pooling and the pixel shuffle layer.
We notice that by adding batch normalisation \cite{ioffe2015batch} before the \textbf{projection convolution step}, the results become more stable. Hence, we use it on all our layers. We denote this layer as a pseudo dilation because it is neither an increasing function nor an extensive function between the input and the output. Nonetheless, it checks these properties between the input and output of the max-pooling.

\subsection{Our Morphological Layers}\label{sec:layerpresentation}

Based on the description of our pseudo morphological dilation, which is our base operator, we describe four more operators in this sub-section: \textbf{pseudo morphological erosion}, \textbf{pseudo morphological pooling}, \textbf{pseudo morphological upsampling,} and \textbf{pseudo morphological gradient}.

The \textbf{pseudo morphological erosion} is constructed with the same exam step as the \textbf{pseudo morphological dilation}, except that instead of doing a max-pooling after the pixel shuffle step, we apply a min-pooling. We propose this operation to check the usefulness of such an operation based on the minimum. We illustrate this layer in Figure \ref{pseudo_mor_shuff}.

Similar to convolution layers with stride, we propose \textbf{pseudo morphological pooling}. This operation consists of first applying a pseudo morphological dilation operation, followed by an extra max-pooling, as shown in Figure \ref{pseudo_mor_shuff}. In other words, by using this approach we achieve the downsampling of the input image.

We propose \textbf{pseudo morphological upsampling}, which is similar to a deconvolution layer, for edge detection \Newcorrection{and semantic segmentation} task\Newcorrection{s}. This operation is implemented by transforming the feature map from $(C, H, W)$ into $(r^2 \times s^2 \times C, H, W)$ through the \textbf{projection convolution step.} Then, we apply the \textbf{pixel shuffle step} to form feature maps with the size $(C,r \times s \times H, r \times s \times W)$. Subsequently, we apply the \textbf{max-pooling step} with stride $r$.

\begin{figure}[htbp]
  \centering
  \includegraphics[width=0.6\textwidth]{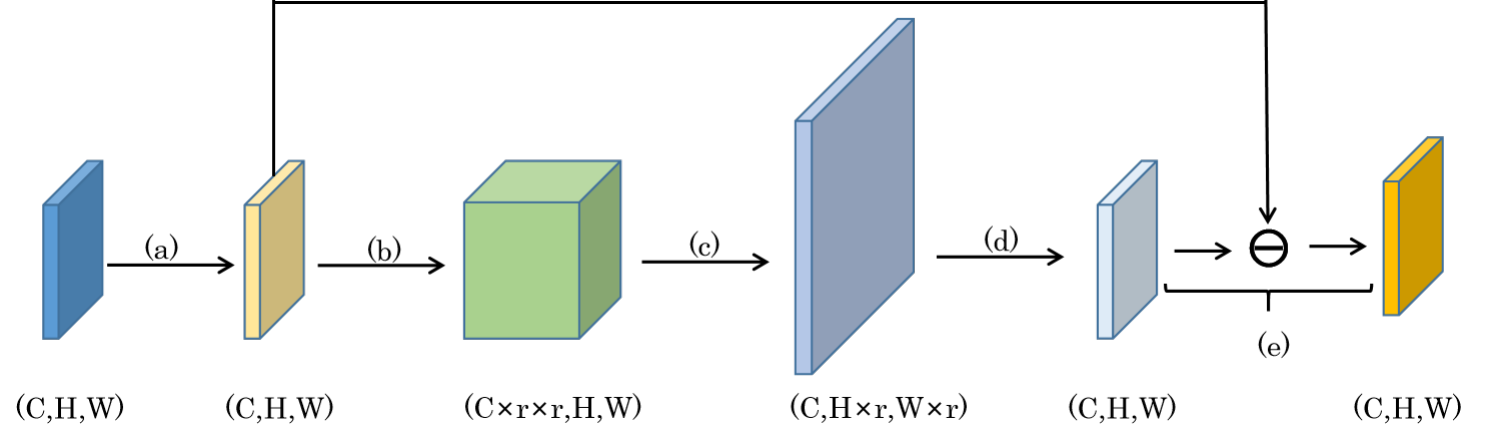} 
  \caption{Overview of the \textbf{pseudo morphological gradient}. The process of (a)(b)(c)(d) is the same as with pseudo morphological dilation. (e) The vector subtraction used to obtain the external pseudo gradient.} 
  \label{pseudo_gradmor_shuff}
\end{figure}

We propose a new layer called the \textbf{pseudo morphological gradient}.  For this layer, we achieve the same \textbf{projection convolution step}, and \textbf{pixel shuffle step}.  Then, we obtain pseudo morphological dilation feature maps using max-pooling, and we obtain the final gradient feature map by performing vector subtraction with the input image. Contrary to the morphological gradient, which  is positive for all pixels, this one is not necessarily positive. This also happens with the morphological Laplacian \cite{serra1988image}, which is of interest for the denoising of images.  Also, the skip connection provided by this gradient layer can help to avoid vanishing gradients. This layer is illustrated in Figure \ref{pseudo_gradmor_shuff}.

\subsection{Neural Architecture Search (NAS)}\label{sec:NAS}

\begin{figure}[htbp]
  \centering
  \includegraphics[width=.55\textwidth]{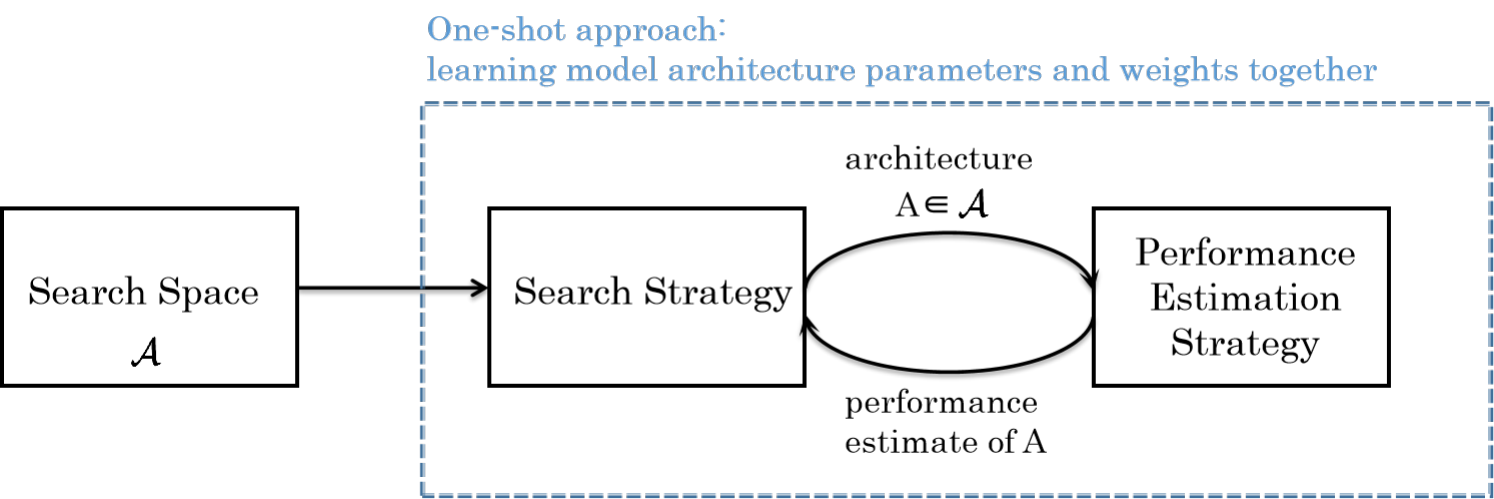} 
  \caption{Components of the Neural Architecture Search (NAS) model} 
  \label{NAS}
\end{figure}

\Gianni{NAS recently identified DNN architectures that exceed human-designed ones in image classification~\cite{liu2019darts,luo2019neural}. As illustrated in Figure \ref{NAS}, NAS needs three key elements:}

\begin{enumerate}
    \item  \textbf{Search Space:} This defines the set of operations (e.g. convolution, fully-connected, pooling) and how we want them to be connected to build valid network architectures. In a \Gianni{sense}, the search space defines the space for the admissible solution. 
    \item \textbf{Search algorithm:} This is the algorithm used to \Gianni{optimize} the architecture
    \item  \textbf{Criterion:} This defines the measure used to estimate or predict the performance of an architecture. 
\end{enumerate}

The \textbf{criterion} used for all the different tasks is the accuracy criterion applied to the validation; we want to optimize this for the given task. The remainder of this section describes our \textbf{search space} and \textbf{search algorithm}.

\subsubsection{Search Space - Cell Search}\label{sec:Cellsearch}

\begin{figure}[htbp]
  \centering
  \includegraphics[width=0.4\textwidth]{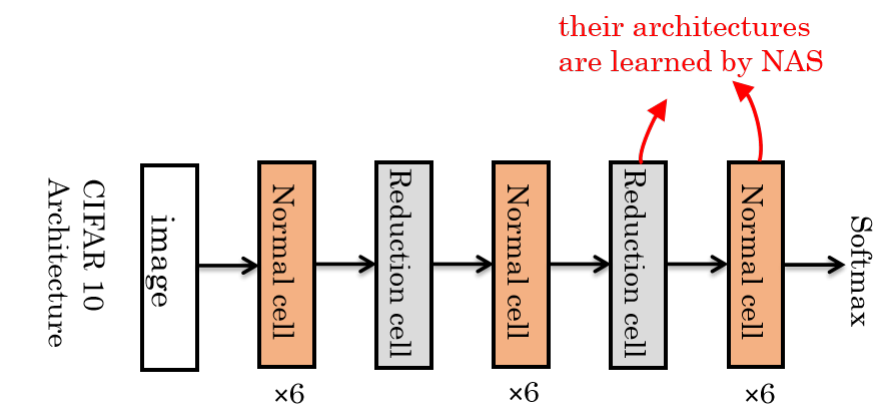} 
  \caption{The overall backbone of a cell-based DNN model for CIFAR 10.} 
  \label{backbone_cifar10}
\end{figure}

In the NAS community, the search space represents the space in which we search the DNN's architecture. %
\Newcorrection{The search space is composed of a set of operations and a backbone explaining how operations can be connected to construct valid network architectures.}
The search space can be classified into two categories. The first one is the
\textbf{Global Search Space}, where the algorithm has to learn all the DNN architecture.
The second one is the \textbf{Cell-Based Search Space}, where the DNN architecture lies in a backbone composed of basic components
called cells, and the goal is to learn what is inside the cells.
Inspired by DNN architecture such as ResNet \cite{resnet} \Correction{or VGG \cite{simonyan2014very}}
, the author of NASNET \cite{zoph2018learning}, who initially proposed Cell-Based Search Space, noted that DNNs are composed of blocks with the same kind of operations that are repeated multiple times. Figure \ref{backbone_cifar10} presents the backbone of a DNN model for CIFAR10 \cite{Krizhevsky09learningmultiple}. 

The Cell-Based Search Space appears to be a more popular alternative than the Global Search Space because the newly discovered neural architecture based on it can be easily transferred between datasets. Two sorts of cells are commonly employed for classification. The first is the standard cell, which preserves the feature map's spatial size, and the second is the reduction cell, which shrinks the feature map's spatial size.

\noindent \textbf{Cell search space for classification}

We tested three cell search spaces for the classification task: two with the morphological layer and one without. The different search spaces are illustrated in Table \ref{table:classificationcell}.

\begin{table}[]
\begin{center}
 \scalebox{0.50}
 {
\resizebox{0.9\columnwidth}{!}
{
\begin{tabular}{|lcl|}
\hline
\multicolumn{1}{|c|}{Cell search space} & \multicolumn{1}{c|}{Cell search space}                             & \multicolumn{1}{c|}{Cell search space}   \\
\multicolumn{1}{|c|}{without morphological layer} & \multicolumn{1}{c|}{with dilation}                             & \multicolumn{1}{c|}{with erosion}                                                 \\ \hline
\multicolumn{1}{|c}{}                             & separable conv $3\times 3$                                     &                                                          \\ \hline
                                                  & separable conv $5\times 5$                                     &                                                               \\ \hline
                                                  & average pooling                                                       &                                                                \\ \hline
                                                  & maximum pooling                                                       &                                                              \\ \hline
\multicolumn{1}{|c|}{$\oslash$}                         & \multicolumn{1}{c|}{pseudo morphological dilation $3\times 3$} & \multicolumn{1}{c|}{pseudo morphological erosion $3\times 3$}   \\ \hline
\end{tabular}
}
}
\caption{Cell search space for the classification task. Separable conv $3\times 3$ and separable conv $5\times 5$ are the separable convolutions introduced in \cite{chollet2017xception}, with kernel sizes of $3\times 3$ and $5\times 5$.}\label{table:classificationcell}
\end{center}
\end{table}

\noindent \textbf{Cell search space for edge detection \Newcorrection{and semantic segmentation}}

For the edge detection \Newcorrection{and semantic segmentation} task\Newcorrection{s}, we \Newcorrection{respectively} propose three cell search spaces and one cell search spaces composed of  six operations. To study the influence of pseudo morphological operations, all the search spaces have the same number of operations. The different search spaces are illustrated in Table \ref{table:edgedetectioncell}.  

\begin{table}[]
\begin{center}
 \scalebox{0.5}
 {
\resizebox{0.9\columnwidth}{!}
{
\begin{tabular}{|lcl|}
\hline
\multicolumn{1}{|c|}{Cell search space} & \multicolumn{1}{c|}{Cell search space}                             & \multicolumn{1}{c|}{Cell search space}   \\
\hline
\multicolumn{1}{|l|}{without morphological layer} & \multicolumn{1}{c|}{with dilation}                             & \multicolumn{1}{c|}{with erosion}                              \\ \hline
\multicolumn{1}{|c}{}                             & cweight $3\times 3$                                            &                                                                \\ \hline
                                                  & separable conv $3\times 3$                                     &                                                                \\ \hline
                                                  & conv $3\times 3$                                               &                                                                \\ \hline
                                                  & average pooling                                                       &                                                                \\ \hline
                                                  & maximum pooling                                                       &                                                                \\ \hline
\multicolumn{1}{|c|}{separable conv $5\times 5$}  & \multicolumn{1}{c|}{pseudo morphological dilation $3\times 3$} & \multicolumn{1}{c|}{pseudo morphological gradient $3\times 3$} \\ \hline
\end{tabular}
}
}
\caption{Cell search space for the  edge detection and semantic segmentation tasks. cweight $3\times3$ is the Squeeze-and-Excitation Network (SENet)
 introduced in \cite{luo2019neural}, with a kernel size of $3\times 3$. }\label{table:edgedetectioncell}
\end{center}
\end{table}

\subsubsection{Search Space of the Architecture}\label{archi_search}

As illustrated in Figure \ref{backbone_cifar10}, the search architecture space is a backbone designed by repeating multiple modules composed of reduction cells and normal cells for classification.  For the classification task, we do not change the search architecture space. However, we propose a search architecture space for \Newcorrection{the segmentation tasks}, where the DNNs play with multiple resolutions, skip connections, and deconvolution.

\noindent \textbf{U-Net architecture search space}
The U-Net architecture \cite{ronneberger2015unet}  inspired our first search architecture space, which we denote U-Net search space.
We start with this search space since U-Net is the state-of-the-art for medical images. 

The U-Net architecture is a fully convolutional network that reinjects the decoder feature map from the encoder information. Hence, the spatial information might be more precise.

Our U-Net search space backbone is similar to the U-Net DNN backbone. However, it is composed of two types of cell: a downsampling segmentation cell and an upsampling segmentation cell, which we denote DownSC and UpSC, respectively. The upsampling segmentation cell is composed of the following operations:

\begin{itemize}
    \item separable conv $3\times 3$     
    \item  separable conv $5\times 5$     
    \item average pooling       
    \item  maximum pooling  
    \item pseudo mophological gradient $3\times3$
    \item  transpose convolution
\end{itemize}

\begin{figure}[htbp]
  \centering
  \includegraphics[width=.6\textwidth]{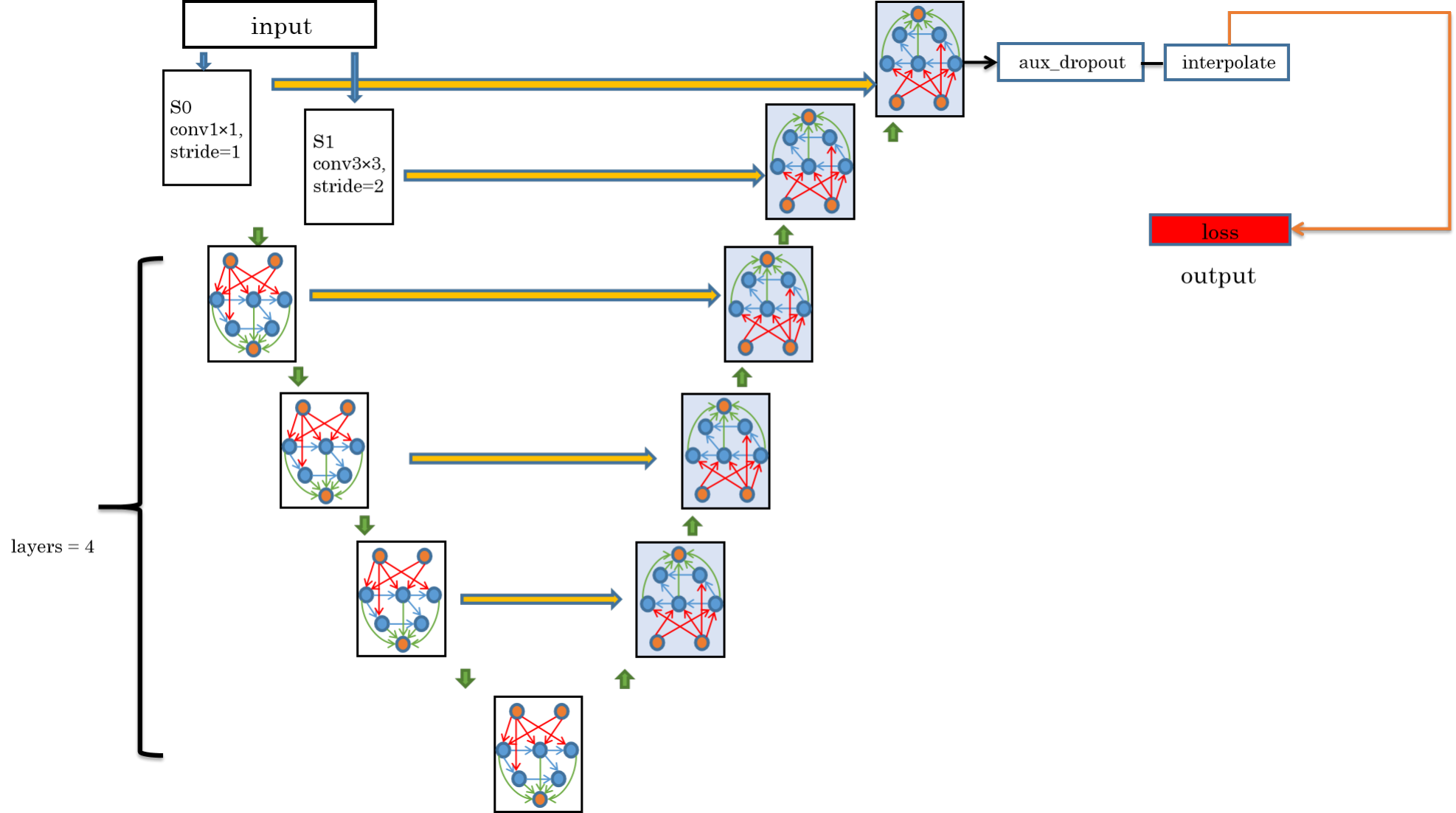} 
  \caption{The overall backbone of the Cell-Based U-net \cite{ronneberger2015unet}\cite{8681706}, where C
equals 8, B represents the number of intermediate nodes, H0 equals 0.25H,
H1 equals 0.5H0, and the rest can be done in the same manner. The
value of W is derived in the same way. In addition, the aux\_dropout is an operation sequential.}
  \label{backbone_unet}
\end{figure}

 The overall  U-Net search backbone is presented in Figure \ref{backbone_unet}.

\noindent \textbf{Multi-scale decoder architecture search space} 

Inspired by Deeplab V3+ \cite{chen2018encoder}, PSPNet \cite{zhao2017pyramid}, and RCF\cite{Liu_2019}, we notice that a good architecture for general images relies on an encoder pretrained on ImageNet\cite{russakovsky2015imagenet} and a decoder that takes multiple resolutions as input and associates them to build the output. Based on that, we propose our new network search space, which  we denote the multi-scale decoder search network space.
\begin{figure}[htbp]
  \centering
  \includegraphics[width=0.6\textwidth]{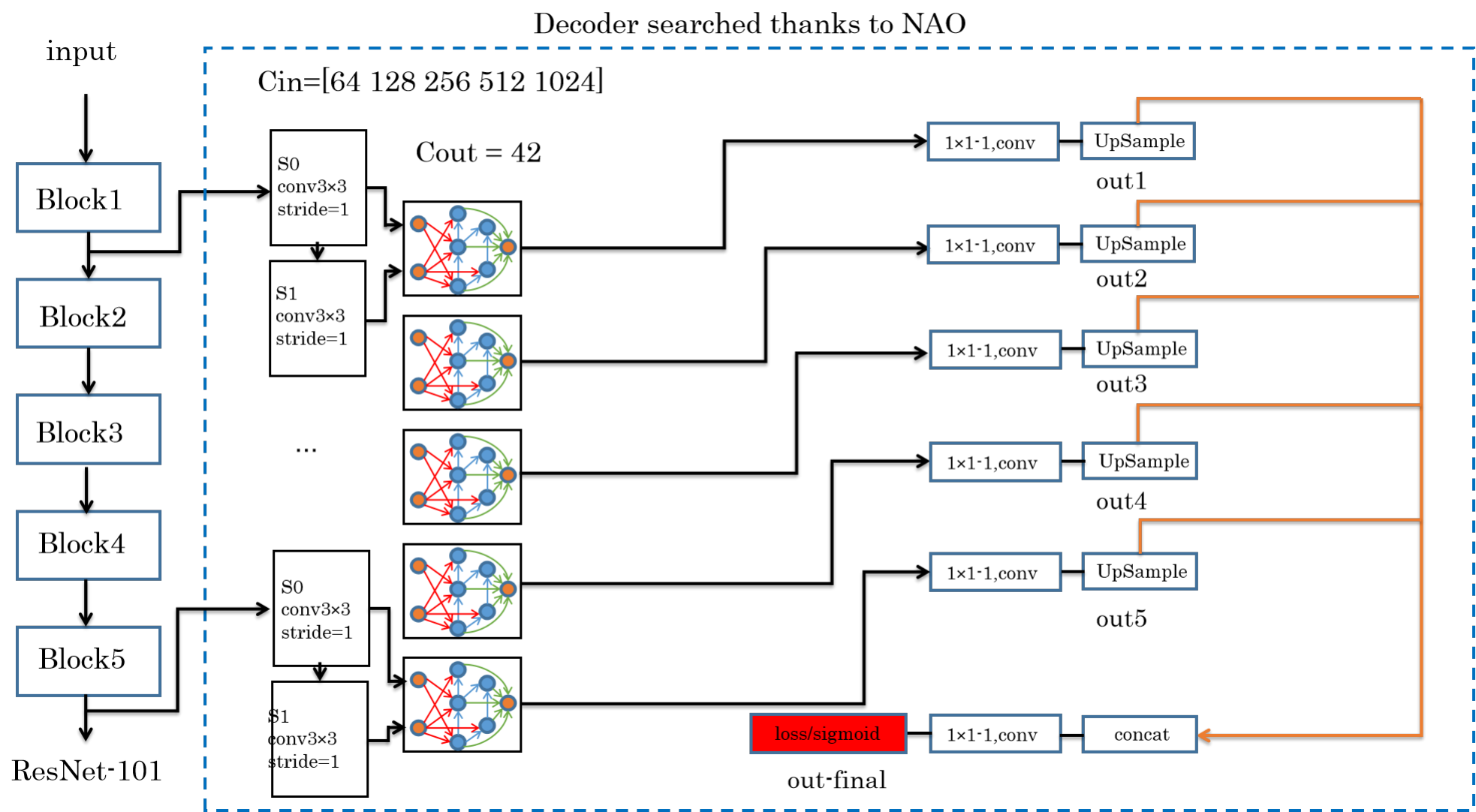} 
  \caption{ResNet101\_Decoder}
  \label{backbone_decoder}
\end{figure}

\Gianni{First, we introduce RCF \cite{Liu_2019}. The RCF algorithm proposes using a classical backbone such as the VGG16 \cite{vggnet} architecture, composed of  13 convolutional layers and three fully connected layers.  The 13 convolutional layers can be subdivided into five blocks. On each block, three convolutions are applied, leading to a feature map for each block. These different feature maps extract information at different resolutions, leading to a multi-scale representation.  This hierarchical information is then merged to form the final output.}

Our new network search space, illustrated in Figure \ref{backbone_decoder}, can be summarized as follows.
We use a ResNet architecture \Correction{\cite{resnet}} pretrained on ImageNet\cite{russakovsky2015imagenet} as an encoder network, similar to \cite{chen2018encoder,zhao2017pyramid, Liu_2019}. ResNet is composed of 4 blocs. We connect each block's output to two layers. The first one is a preprocessing layer that we denote $S_0$. This layer is composed of a $3 \times 3 $ convolution and outputs 42 feature maps. This operation allows us to control the depth map that will enter the cells.
 The second preprocessing layer, applied on the output of $S_0$ 
 and denoted  $S_1$, is composed of a $3 \times 3 $ convolution and outputs 42 feature maps. The inputs of the cells are  the outputs of $S_0$ and $S_1$, so that the cell learns if it wants to use $S_0$ and/or $S_1$.
After each block,  we use a $1\times 1$ convolution on the cells' output to reduce the channel's number to one. This convolution is followed by an upsampling to resize all the feature maps to input size, and we concatenate all the upsampled results. Finally, we use a $1\times1$ to produce the final edge detection map.

\noindent \Newcorrection{\textbf{DeeplabV3+  architecture search space}}

\begin{figure}[htbp]
  \centering
  \includegraphics[width=0.6\textwidth]{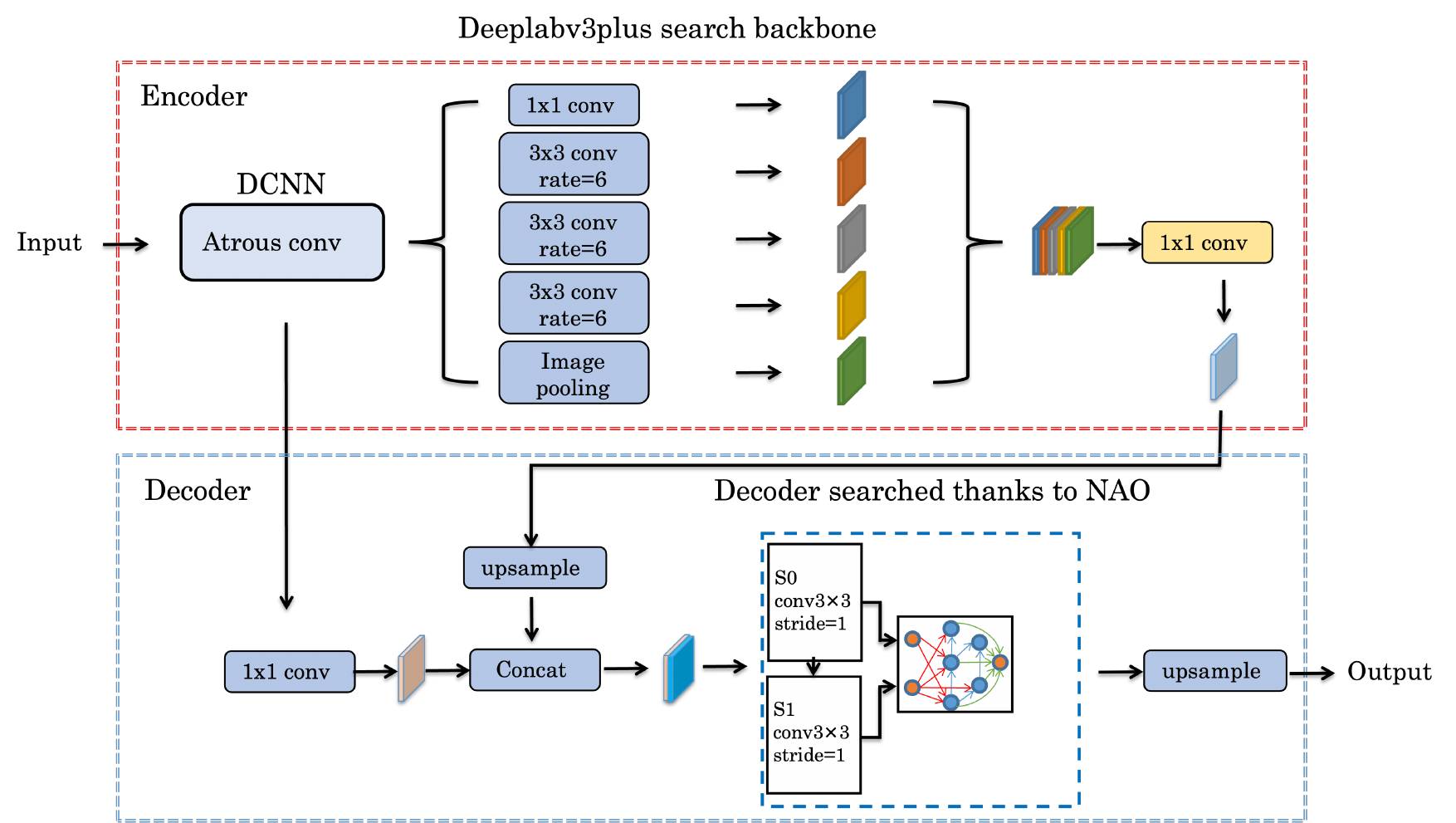} 
  \caption{\Newcorrection{DeeplabV3+ architecture search space}}
  \label{DeeplabV3}
\end{figure}

\Newcorrection{Similarly to Deeplab v3+ \cite{chen2018encoder}, our new network search space, illustrated in Figure \ref{DeeplabV3}, can be summarized as follows.
It is based on an Atrous Spatial Pyramid Pooling (ASPP), which is able to encode multi-scale contextual information. After having concatenated the output of the ASPP with a low-level feature like Deeplab v3+, we replace the $3\times 3$ convolution by a cell wherein we search for the operations.}

\subsubsection{Neural Architecture Optimization (NAO)}\label{sec:NAO}

NAO~\cite{luo2019neural} is an optimization algorithm that searches for the best architectures based on the following principle. First, NAO is a \textbf{Cell-Based Search Space} algorithm, like most modern NAS. Hence, we merely need to learn the adjacency matrix that represents the cells. Secondly, NAO is a \textbf{two-step algorithm} that first searches for the best architecture in \textit{step 1}. Then, in \textit{step 2}, with this architecture, NAO optimizes the DNN's weights to search for the best model. Thirdly, 
NAO does not directly optimize the cell parameters, and it performs its optimization on a \textbf{latent space} of cell parameters.

\begin{figure}[htbp]
  \centering
  \includegraphics[width=.6\textwidth]{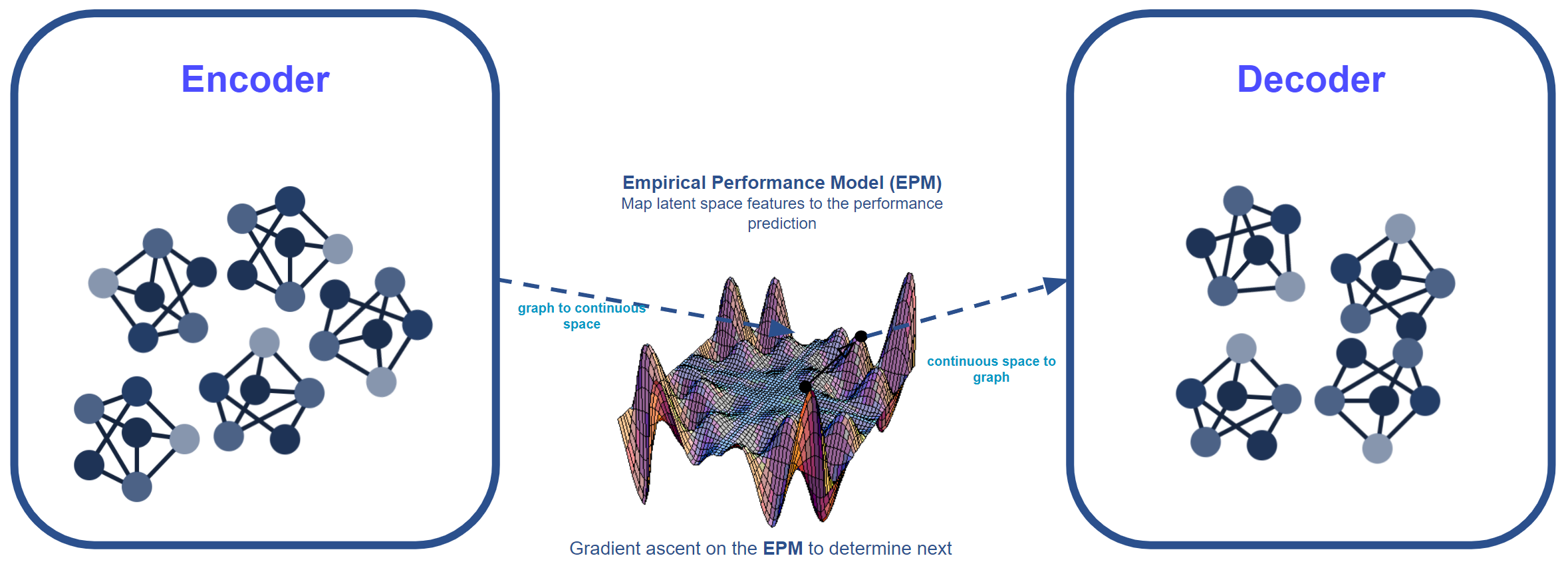} 
  \caption{Overview of the NAO algorithm. The encoder maps architecture $x$
onto a continuous space represented by $e_{x}$. Afterwards, the predictor
optimizes $e_{x}$ by maximizing the output of performance using gradient
descent. The output of the predictor is represented as $e_{x^{\prime}}$. The decoder transforms $e_{x^{\prime}}$ into a new discrete architecture $x^{\prime}$\cite{luo2019neural}}
  \label{NAO_algo}
\end{figure}

The NAO process is illustrated in Figure \ref{NAO_algo}.
In detail, the NAO algorithm consists of an encoder, a predictor, and a decoder network\cite{luo2019neural}.
The encoder of NAO \cite{luo2019neural} takes as input a randomly generated architecture sequence describing an architecture, and then maps it onto a continuous space C. Specifically, the encoder is denoted as $\mathrm{E}: \mathrm{X} \rightarrow \mathrm{C}$. Let us write $e_{x}=E(x)$ as the latent representation of the DNN architecture.

The performance predictor \cite{luo2019neural} $\mathrm{P}: \mathrm{C} \rightarrow \mathrm{R}$ maps the latent representation of an architecture x onto its performance $s_{x}$. With an architecture $x$ and its performance $s_{x}$ as training data, the optimization of $P$ aims to minimize the least-square regression $\text {loss }=\left(s_{x}-P(E(x))\right)^{2}$.

The decoder of NAO \cite{luo2019neural}, which is similar to the decoder in the DNN model, is responsible for decoding out the string tokens in $x$, taking $e_{x}$ as input. Mathematically, the decoder is denoted as function $\mathrm{D}: \mathrm{C} \rightarrow \mathrm{X}$, which decodes the input taking $e_{x}$.
The training process consists of optimizing the following loss: $\sum_{x \in X} \log P_{D}(x \mid E(x))$.

The encodee-decoder learns to build a latent space that can represent the space of the architecture. The performance predictor learns to map this space onto its performance for the given task. Finally, a new architecture is generated by trying to determine which one has the best performance.

\Correction{
In this work, NAO was our preferred NAS method for the following three \Newcorrection{reasons}:
\begin{itemize}
    \item NAO incorporates recent search strategies to reduce the computational cost (a Cell-Based Search Space and weight sharing).
    \item NAO can be executed with a small computational time. 
    \item NAO allows the easy customization of the backbone architecture.
\end{itemize}
}

\section{Experiments}

In this section, we explain the experiments confirming our morphological layers' utility.
We consider three different types of experiments: classification, edge detection and semantic segmentation.

\subsection{Classification task}

\begin{figure}[tbp]
\centering
    \subfloat[Normal cell for CIFAR-10 with \textbf{pseudo morphological dilation}]{\includegraphics[width=0.30\linewidth]{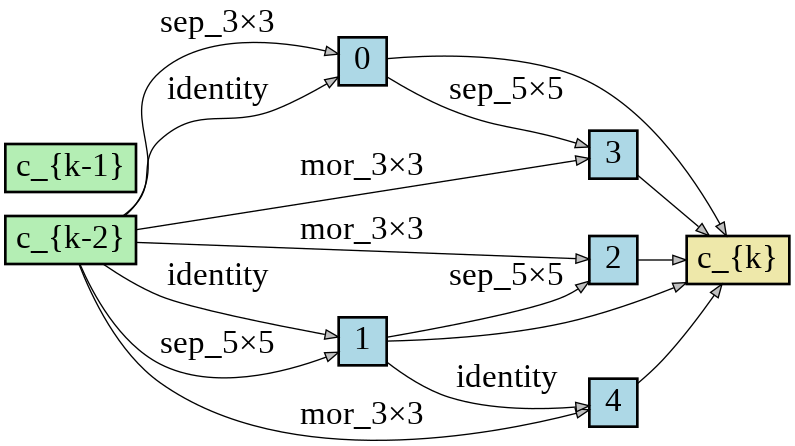}\label{normal_CIFAR10_1}}
    \subfloat[Reduction cell for CIFAR-10 with \textbf{pseudo morphological dilation}]{\includegraphics[width=0.30\linewidth]{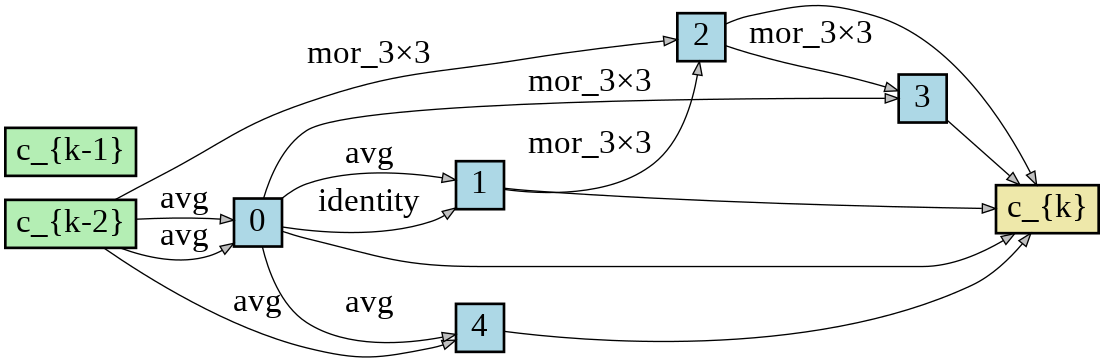}\label{reduction_CIFAR10}}\\
    \subfloat[\Correction{Adjacency matrix of the} Normal cell for CIFAR-10 with \textbf{pseudo morphological dilation} ]{\includegraphics[width=0.50\linewidth]{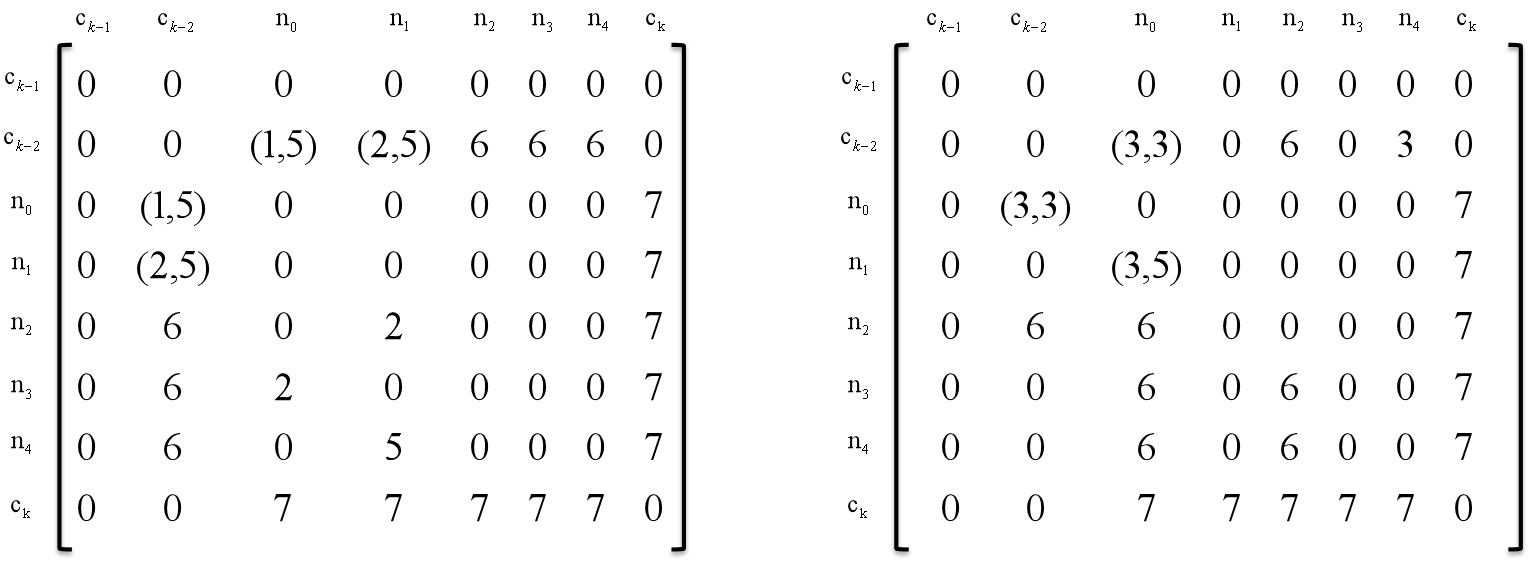}\label{normal_CIFAR10_2}}
\caption{\Gianni{(a) and (b) represent the} cell structure for the classification of CIFAR10, where $c_{k-1}$ represents the previous output and $c_{k-2}$ represents the output of the cell just before $c_{k-1}$. The blue numbers 0-4 represent the intermediate nodes, and every edge represents different operations. In addition, $c_{k}$ represents the output of the current cell, which concatenates the results of the intermediate node 0-4. \Gianni{(c) represents} the adjacency matrix for the normal cell and the reduction cell. From left to right: the adjacency matrix of the normal cell, the adjacency matrix of the reduction cell. Among them, the numbers 1,2,3,4,5,6, and 7 respectively represent the following layers: separable conv $3\times3$, separable conv $5\times5$, avg pool, max pool, identity, pseudo morphological dilation $3\times3$, and concatenate operation.}\label{cell_CIFAR10_Structure}
\end{figure}

We evaluate our layers on two datasets, CIFAR10\cite{Krizhevsky09learningmultiple}  and CIFAR100\cite{Krizhevsky09learningmultiple},
which are composed of 50 000 RGB training images and 10 000 RGB test images of the size $32 \times 32$. CIFAR10 has 10 classes, while CIFAR100 has 100 classes.
To train the DNN, we used the cross-entropy loss and reported the classification error, and we used the same experimental protocol as NAO uses for CIFAR 10. %
During the architecture search, we use a small network with $B=5$ (number of nodes), $N=3$ (number of normal cells), and $F=8$ (number of channels), and search the architecture for four iterations; each of them is composed of 50 epochs. After the best cell architectures are found, we increase the architecture with $B=5$, $N=6$, $F=36$ and optimize the weight of the DNNs for 600 epochs. The batch size for the two-step is 128.
 The results in Table \ref{performance_on_cifar10} are the mean of 3 seeds. The mean error of our architecture with the morphological layer is 2.65\%, which is below the 2.93\% of NAO \cite{luo2019neural}. After attempting to replace the morphological dilation with erosion, we see that the result decreases.
Hence, we noted on CIFAR 10 that the morphological dilation layer improves the performance of the DNN. We also note from Figure \ref{cell_CIFAR10_Structure} that NAO uses this operation since it is present in 3  out of the 15 edges for the normal cell and 4 out of the 15 edges for the reduction cell. Morphological operations seem to bring information not presented by traditional convolutional layers, which can help to improve performance. 
\Correction{Moreover, it seems that the training with the BN has a positive effect since it enables the performance to improve from 2.84 to 2.65}

Figure \ref{cell_CIFAR10_Structure} illustrates the normal and reduction cell structures that we learned for the CIFAR10 \cite{Krizhevsky09learningmultiple} classification task. The cells are represented by a graph, which can be characterized by an adjacency matrix as drawn in Figure \ref{cell_CIFAR10_Structure}.

We also train an architecture search on CIFAR100 and check that we can confirm our previous result. Table \ref{performance_on_cifar100} presents our results for CIFAR100.
We note that the morphological layer seems to improve the performance of the classification task. Furthermore, training from scratch on the architecture search on CIFAR100 brings worse results than transferring the architecture learned on CIFAR10.

\begin{table}%
\centering
 \scalebox{0.80}
 {
\begin{tabular}{c c c c}
\toprule
Method&B&Error(\%)&\Newcorrection{GPU Days}\\
\midrule 
\midrule 
ResNet with stochastic depth\cite{huang2016deep}&&5.25&\Newcorrection{0.63}\\
Wide ResNet\cite{zagoruyko2017wide}&&4.00&\Newcorrection{/}\\

ENAS+Cutout\cite{pham2018efficient}&$5$&3.54&\Newcorrection{0.45}\\
Block-QNN-S more filters\cite{zhong2018practical}&&3.54&\Newcorrection{3}\\
DenseNet-BC\cite{huang2018densely}&&3.46&\Newcorrection{/}\\
PNAS+Cutout\cite{liu2018progressive}&$5$&3.41&\Newcorrection{225}\\
DARTS+Cutout\cite{liu2019darts}&$5$&$2.83$&\Newcorrection{4}\\
\Newcorrection{MergeNAS(2nd-order)+cutout\cite{MergeNAS}}&& \Newcorrection{$2.68$}&\Newcorrection{0.6}\\
\Newcorrection{PC-DARTS+cutout\cite{pcDarts}}&&\Newcorrection{$2.57$}&\Newcorrection{0.1}\\
\Newcorrection{few-shot DARTS-Small+cutout\cite{few}}&&\Newcorrection{$2.31$}&\Newcorrection{1.35}\\
\Newcorrection{One-Stage ISTA\cite{ISTA}}&&\Newcorrection{$2.36$}&\Newcorrection{2.3}\\
\midrule 
NAONet-WS\cite{luo2019neural}&$5$&$3.53$&\Newcorrection{0.3}\\
NAONet-WS+Cutout\cite{luo2019neural}&$5$&$2.93$&\Newcorrection{0.3}\\
\midrule 
\Correction{NAONet-WS+Cutout+}\\\Correction{\textbf{pseudo morphological dilation without BN}}&$5$&\Correction{$\textbf{2.84}$}&\Newcorrection{0.3}\\
NAONet-WS+Cutout+\\\textbf{pseudo morphological dilation}&$5$&$\textbf{2.65}$&\Newcorrection{0.3}\\
NAONet-WS+Cutout+\\\textbf{pseudo morphological erosion}&$5$&$3.20$&\Newcorrection{0.3}\\
\bottomrule
\end{tabular}
}
\caption{CIFAR10 Performance of NAONet with  pseudo morphological operations, where B represents the number of nodes within the cell. Error represents the accuracy error.}\label{performance_on_cifar10}
\end{table}

\begin{table}%
\centering
 \scalebox{0.80}
 {
\begin{tabular}{c c c}
\toprule
Method&B&Error(\%)\\
\midrule 
\midrule 

ResNet with stochastic depth\cite{huang2016deep}&&24.98\\
Wide ResNet\cite{zagoruyko2017wide}&&19.25\\
Block-QNN-S more filters\cite{zhong2018practical}&&18.06\\
PNAS+Cutout$^*$\cite{liu2018progressive}&$5$&$17.44$\\
DenseNet-BC\cite{huang2018densely}&&17.18\\
ENAS+Cutout$^*$\cite{pham2018efficient}&$5$&$16.44$\\
\Newcorrection{One-Stage ISTA\cite{ISTA}}&&\Newcorrection{$16.9$}\\
\Newcorrection{DHA\cite{DHA}}&&\Newcorrection{$16.07$}\\
\midrule 
NAONet-WS+Cutout$^*$\cite{luo2019neural}&$5$&\textbf{15.67}\\
\midrule 
NAONet-WS+Cutout+\\\textbf{ without pseudo morphological dilation}&$5$&16.9\\
NAONet-WS+Cutout+\\\textbf{pseudo morphological dilation}&$5$&\textbf{16.23}\\
\bottomrule
\end{tabular}
}
\caption{CIFAR100 Performance of NAONet with  pseudo morphological operations, where B represents the number of nodes within the cell. Error represents the accuracy error. Please note that the results of the techniques with a star $^*$ train the architecture search on CIFAR10 and learn the model on CIFAR 100.}\label{performance_on_cifar100}
\end{table}

\subsection{Edge detection task}

We \Newcorrection{propose} a method for extracting image edges to highlight the power of morphological operators in deep learning frameworks.
We recommended two backbones for this task: the U-Net search backbone and the multi-scale decoder architecture search space. These are detailed in section \ref{archi_search}.

We train our DNNs on BSDS500 \cite{amfm_pami2011}, which comprises 200 training, 100 validation, and 200 test images.  Up to 9 annotators labeled each image. Like previous works \cite{Liu_2019,liu2016learning,yang2016object,kokkinos2015pushing},  we use the training set and validation set for tuning the DNN and the test set for the evaluation and mix the augmented training data of BSDS500 with the flippedVOC Context dataset \cite{mottaghi2014role}.

On medical images, the U-Net backbone offers state-of-the-art performance for semantic segmentation,  yet as illustrated in Table~\ref{performance_grad}, this backbone does not give good results for edge detection.
Subsequently, we propose a multi-scale decoder architecture search space that we denote \textbf{NAO-Multi-scale}. This backbone is inspired by 
traditional algorithms \cite{WEN201884} that perform well on this task. This algorithm learns a decoder using an encoder pre-trained on ImageNet \cite{russakovsky2015imagenet}.

To evaluate our novel algorithm, we use the F1-score, which is the harmonic average of the precision and recall and ranges between 0 and 1, with higher values being better. We evaluate the F1 score for each image of the test set of BSDS500 \cite{amfm_pami2011} at different thresholds for the edge prediction. We apply different thresholds since our results are edge probabilities with values between zero and one. The closer to one the edge value is, the more likely this edge is to be correct. However, this metric does not provide a result for the entire dataset. Hence, similar to \cite{Liu_2019,liu2016learning,yang2016object,kokkinos2015pushing}, the Optimal Dataset Scale  (ODS) and Optimal Image Scale (OIS) are used to provide a metric for the whole dataset.

The Optimal Dataset Scale (ODS), where one chooses the optimal threshold for the entire dataset before applying the F1 score, and the Optimal Image Scale (OIS),  where one chooses the optimal threshold per-image before using the F1 score, are two metrics for evaluating the quality of the edge detection algorithm on the whole dataset. 
The OIS is always slightly better than the ODS since it considers the best scale for each image. The OIS corresponds to the optimistic situation where we have the optimal threshold for each image of the dataset. For more information about these classical measures for edge
detection, refer to \cite{amfm_pami2011}\cite{8516362}.

\begin{figure}[htpb]
  \centering
        \subfloat[image]{
        \begin{minipage}[b]{0.12\textwidth}
        \includegraphics[width=0.8\linewidth]{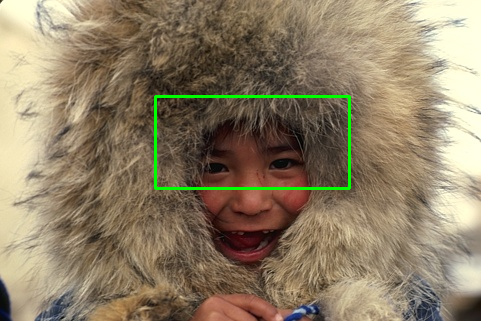}\vspace{4pt}
        \includegraphics[width=0.8\linewidth]{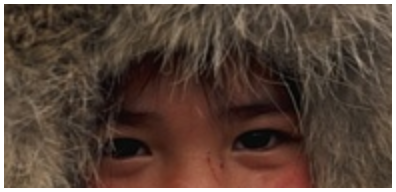}
        \end{minipage}}
        \subfloat[ground \\truth]{
        \begin{minipage}[b]{0.12\textwidth}
        \includegraphics[width=0.8\linewidth]{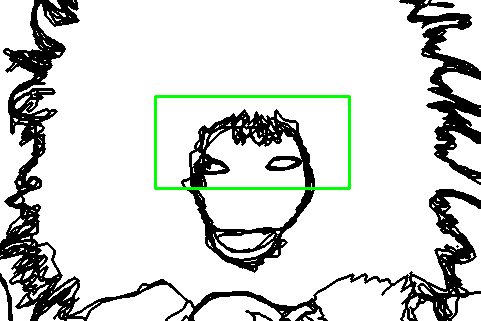}\vspace{4pt}
        \includegraphics[width=0.8\linewidth]{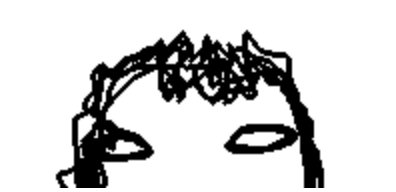}
        \end{minipage}}
        \subfloat[F1(RCF)\\=0.788]{
        \begin{minipage}[b]{0.12\textwidth}
        \includegraphics[width=0.8\linewidth]{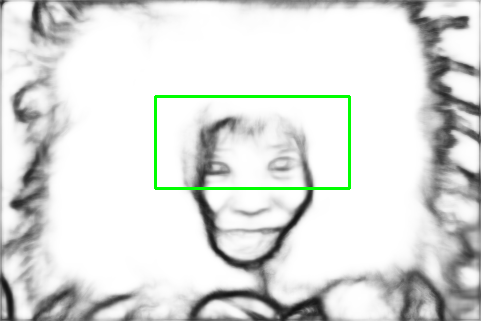}\vspace{4pt}
        \includegraphics[width=0.8\linewidth]{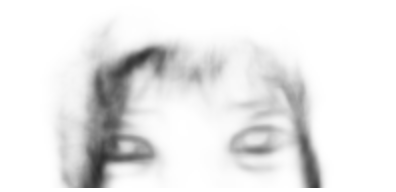}
        \end{minipage}}
        \subfloat[F1(ours)\\=0.824]{
        \begin{minipage}[b]{0.12\textwidth}
        \includegraphics[width=0.8\linewidth]{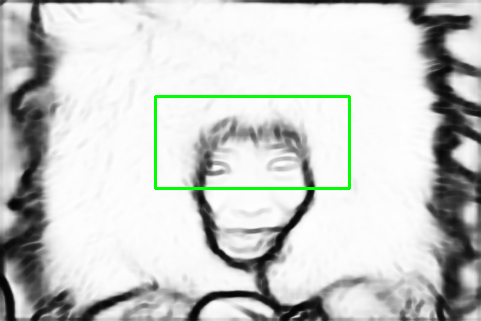}\vspace{4pt}
        \includegraphics[width=0.8\linewidth]{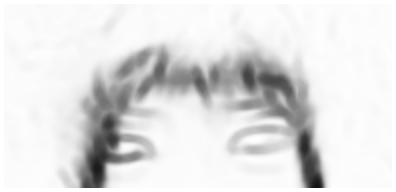}
        \end{minipage}}
        \subfloat[image]{
        \begin{minipage}[b]{0.12\textwidth}
        \includegraphics[width=0.8\linewidth]{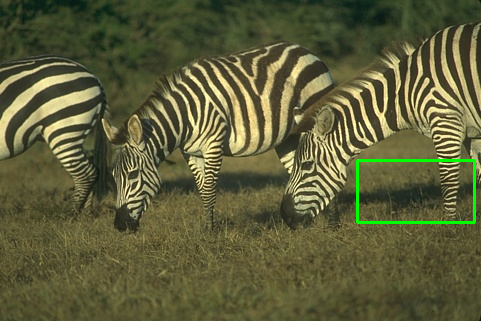}\vspace{4pt}
        \includegraphics[width=0.8\linewidth]{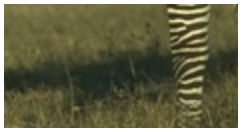}
        \end{minipage}}
        \subfloat[ ground \\truth]{
        \begin{minipage}[b]{0.12\textwidth}
        \includegraphics[width=0.8\linewidth]{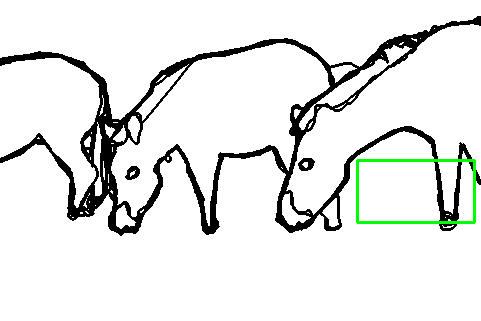}\vspace{4pt}
        \includegraphics[width=0.8\linewidth]{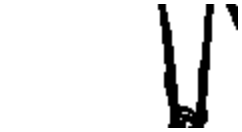}
        \end{minipage}}
        \subfloat[F1(RCF)\\=0.845]{
        \begin{minipage}[b]{0.12\textwidth}
        \includegraphics[width=0.8\linewidth]{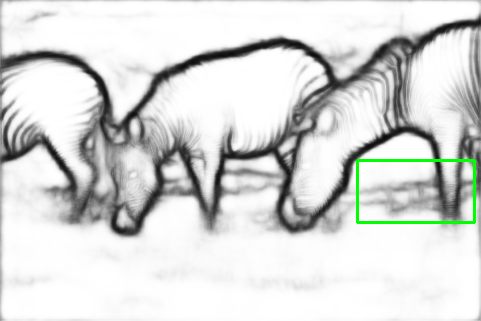}\vspace{4pt}
        \includegraphics[width=0.8\linewidth]{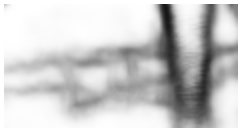}
        \end{minipage}}
        \subfloat[F1(ours)\\=0.900]{
        \begin{minipage}[b]{0.12\textwidth}
        \includegraphics[width=0.8\linewidth]{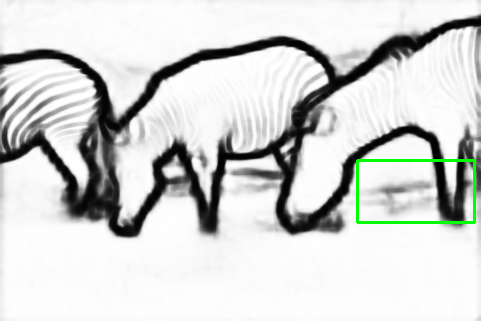}\vspace{4pt}
        \includegraphics[width=0.8\linewidth]{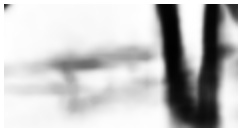}
        \end{minipage}}
        
        \Newcorrection{
        }

  \caption{Overview of the results of different methods on BSD500 \cite{amfm_pami2011}. From left to right: original image, \small{ground truth},  RCF \cite{Liu_2019} predicted edge map, and NAO\_multi\_scale predicted edge map.
We can see in these examples that RCF has more false negatives than  NAO\_multi\_scale. }
  \label{BSD500_FN0}
\end{figure}

\begin{figure}[h]
  \centering
  
    \subfloat[image]{
        \begin{minipage}[b]{0.12\textwidth}
        \includegraphics[width=0.8\linewidth]{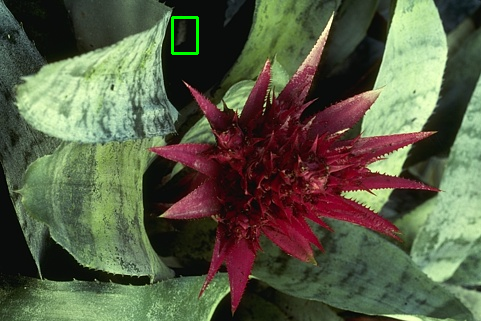}\vspace{4pt}
        \includegraphics[width=0.8\linewidth]{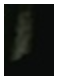}
        \end{minipage}}
        \subfloat[ground \\truth]{
        \begin{minipage}[b]{0.12\textwidth}
        \includegraphics[width=0.8\linewidth]{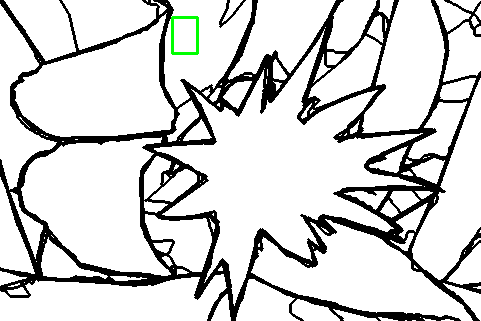}\vspace{4pt}
        \includegraphics[width=0.8\linewidth]{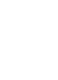}
        \end{minipage}}
        \subfloat[F1(RCF)\\=0.920]{
        \begin{minipage}[b]{0.12\textwidth}
        \includegraphics[width=0.8\linewidth]{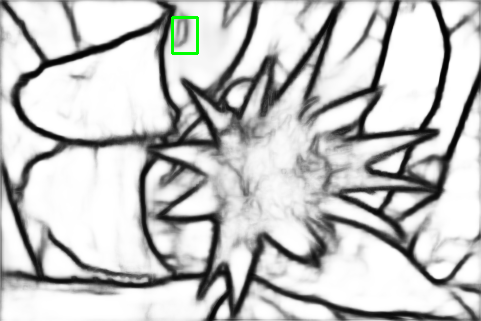}\vspace{4pt}
        \includegraphics[width=0.8\linewidth]{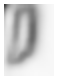}
        \end{minipage}}
        \subfloat[F1(ours)\\=0.933]{
        \begin{minipage}[b]{0.12\textwidth}
        \includegraphics[width=0.8\linewidth]{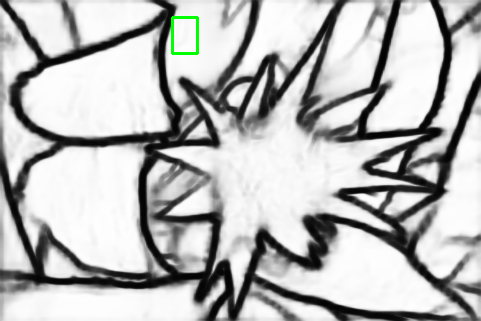}\vspace{4pt}
        \includegraphics[width=0.8\linewidth]{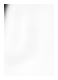}
        \end{minipage}}
  \subfloat[image]{
        \begin{minipage}[b]{0.12\textwidth}
        \includegraphics[width=0.8\linewidth]{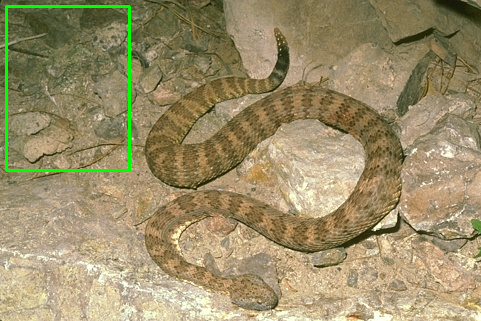}\vspace{4pt}
        \includegraphics[width=0.8\linewidth]{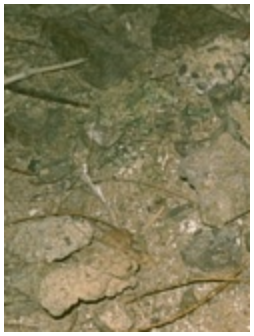}
        \end{minipage}}
        \subfloat[ground \\truth]{
        \begin{minipage}[b]{0.12\textwidth}
        \includegraphics[width=0.8\linewidth]{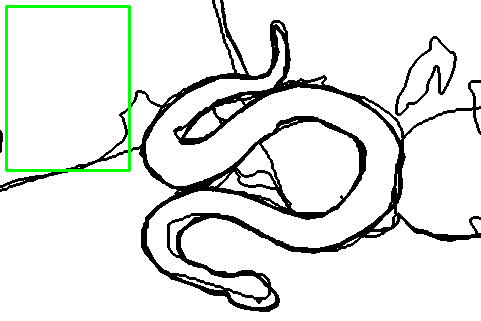}\vspace{4pt}
        \includegraphics[width=0.8\linewidth]{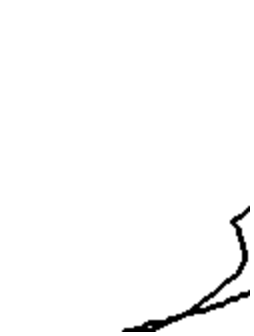}
        \end{minipage}}
        \subfloat[F1(RCF)\\=0.808]{
        \begin{minipage}[b]{0.12\textwidth}
        \includegraphics[width=0.8\linewidth]{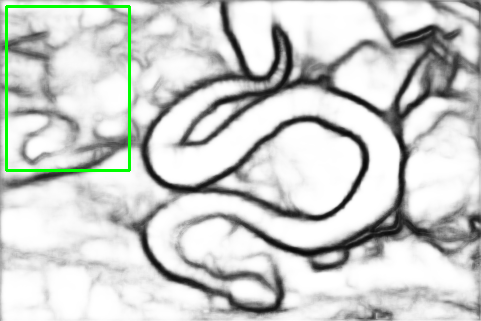}\vspace{4pt}
        \includegraphics[width=0.8\linewidth]{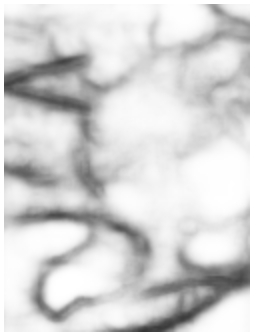}
        \end{minipage}}
        \subfloat[F1(ours)\\=0.833]{
        \begin{minipage}[b]{0.12\textwidth}
        \includegraphics[width=0.8\linewidth]{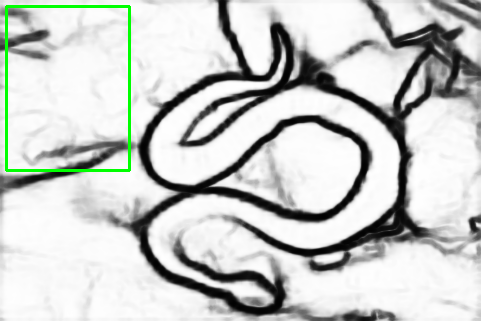}\vspace{4pt}
        \includegraphics[width=0.8\linewidth]{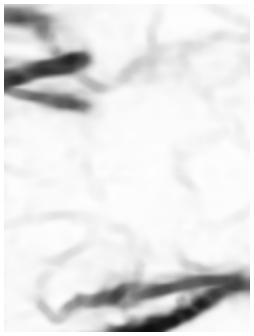}
        \end{minipage}}

  \caption{ Overview of the results of different methods on BSD500 \cite{amfm_pami2011}. From left to right: original image, \small{ground truth},  RCF \cite{Liu_2019} predicted edge map, and NAO\_multi\_scale predicted edge map.
We can see in these examples that RCF has more false positives than  NAO\_multi\_scale. }
  \label{BSD500_FP}
\end{figure}

\begin{figure}[h]
  \centering
  \Newcorrection{
        }
        \subfloat[image]{
        \begin{minipage}[b]{0.12\textwidth}
        \includegraphics[width=0.8\linewidth]{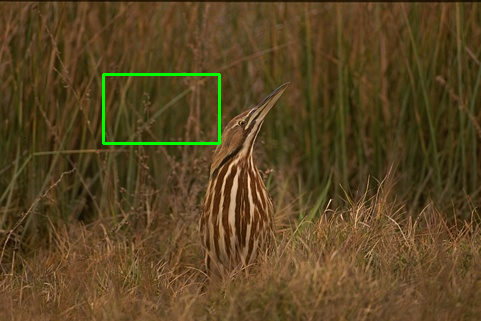}\vspace{4pt}
        \includegraphics[width=0.8\linewidth]{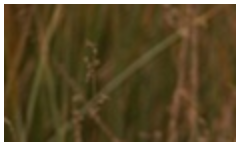}
        \end{minipage}}
        \subfloat[ground \\truth]{
        \begin{minipage}[b]{0.12\textwidth}
        \includegraphics[width=0.8\linewidth]{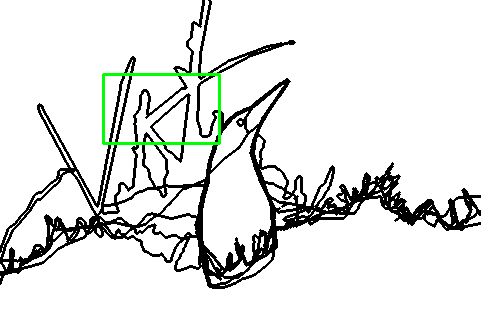}\vspace{4pt}
        \includegraphics[width=0.8\linewidth]{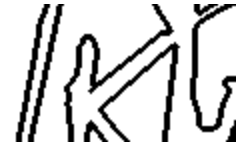}
        \end{minipage}}
        \subfloat[F1(RCF)\\=0.733]{
        \begin{minipage}[b]{0.12\textwidth}
        \includegraphics[width=0.8\linewidth]{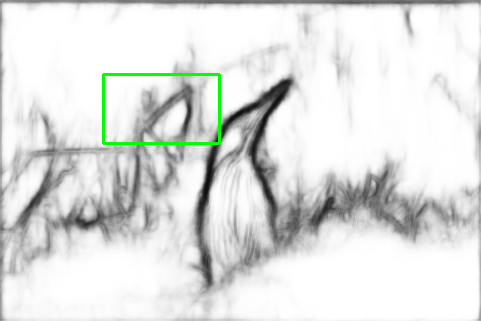}\vspace{4pt}
        \includegraphics[width=0.8\linewidth]{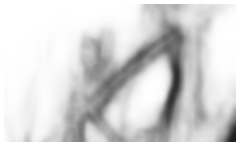}
        \end{minipage}}
        \subfloat[F1(ours)\\=0.722]{
        \begin{minipage}[b]{0.12\textwidth}
        \includegraphics[width=0.8\linewidth]{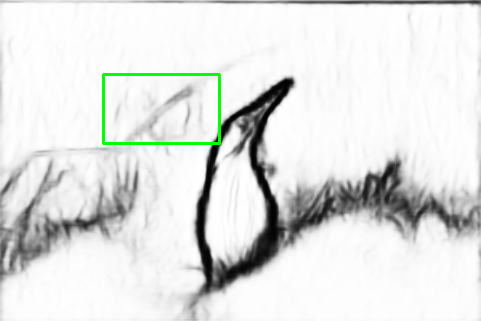}\vspace{4pt}
        \includegraphics[width=0.8\linewidth]{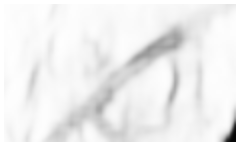}
        \end{minipage}}
        \subfloat[image]{
        \begin{minipage}[b]{0.12\textwidth}
        \includegraphics[width=0.8\linewidth]{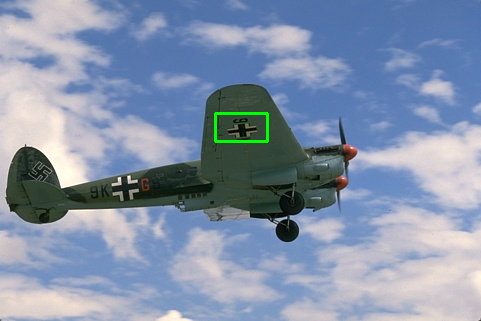}\vspace{4pt}
        \includegraphics[width=0.8\linewidth]{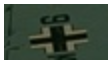}
        \end{minipage}}
        \subfloat[ground \\truth]{
        \begin{minipage}[b]{0.12\textwidth}
        \includegraphics[width=0.8\linewidth]{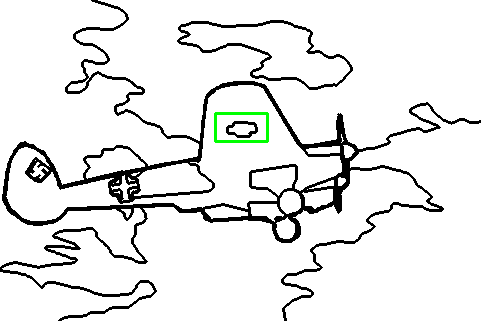}\vspace{4pt}
        \includegraphics[width=0.8\linewidth]{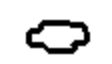}
        \end{minipage}}
        \subfloat[F1(RCF)\\=0.799]{
        \begin{minipage}[b]{0.12\textwidth}
        \includegraphics[width=0.8\linewidth]{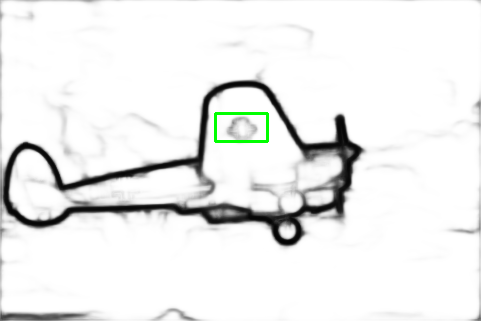}\vspace{4pt}
        \includegraphics[width=0.8\linewidth]{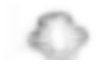}
        \end{minipage}}
        \subfloat[F1(ours)\\=0.787]{
        \begin{minipage}[b]{0.12\textwidth}
        \includegraphics[width=0.8\linewidth]{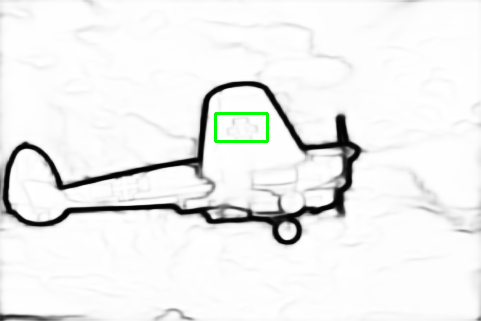}\vspace{4pt}
        \includegraphics[width=0.8\linewidth]{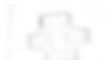}
        \end{minipage}}
  \centering
  \caption{ Overview of the results of different methods on BSD500 \cite{amfm_pami2011}. From left to right: original image, \small{ground truth},  RCF\cite{Liu_2019} predicted edge map, and NAO\_multi\_scale predicted edge map.
We can see in these examples that NAO\_multi\_scale has more false \Newcorrection{negatives} than RCF. }
  \label{BSD500_Bad}
\end{figure}

Figure \ref{BSD500_FN0} presents examples of where RCF is unable to detect edges, while Figure \ref{BSD500_FP} offers examples of where RCF detects edges where there are no edges. Finally, Figure \ref{BSD500_Bad} gives some examples of where our algorithm fails.

\begin{table}[htbp]
\centering
 \scalebox{0.80}
 {
\begin{tabular}{c c c c c}
\toprule
Method&ODS&OIS&AP&R50\\
\midrule 
\midrule 
DeepEdge \cite{bertasius2015deepedge} &0.753 &0.772& 0.807 & ~ \\
$N^4$-Fields \cite{ganin2014} &0.753 &0.767& 0.780 & ~ \\
HFL \cite{bertasius2015highforlow} &0.767 &0.788& 0.800 & ~ \\
HED \cite{xie2015holisticallynested} &0.782 &0.804& 0.833 & ~ \\
RDS \cite{liu2016learning} &0.792 &0.810& ~ & ~ \\
CEDN \cite{yang2016object}  & 0.788 &0.804 & ~ & ~ \\
AMH-Net(fusion) \cite{xu2018learning}  & 0.798 &0.829 & 0.869 & ~ \\
CED \cite{Wang_2019} &0.803 &0.820& 0.871 & ~ \\
MIL+G-DSN+VOC+MS+NCuts \cite{kokkinos2015pushing} &0.813 & \textbf{0.831} & ~ & ~ \\
RCF\_ResNet101 \cite{Liu_2019}& 0.812 & 0.829 & &\\
\Newcorrection{RCN-VOC-1\cite{RCN}}&\Newcorrection{0.812}&\Newcorrection{0.827}&\Newcorrection{0.822}\\
\Newcorrection{CATS-BDCN\cite{cats_BDCN}}&\Newcorrection{0.812}&\Newcorrection{0.828}&&\\

\midrule 
\textbf{NAO-U-NET(ours)} &0.788 &0.808 &0.814& 0.899\\
with dilation search space\\
\midrule 

\textbf{NAO-Multi-scale(ours)} without morphological search space&0.812 $\pm0.001$ &0.830 $\pm0.001$ &0.827$\pm0.014$  &0.903$\pm0.00471$\\
\textbf{NAO-Multi-scale(ours)} with dilation search space &0.809 $\pm0.001$& 0.829 $\pm0.001$& 0.825 $\pm0.0149$& 0.900 $\pm0.0048$\\
\textbf{NAO-Multi-scale(ours)}  with gradient search space&\textbf{0.814} $\pm0.001$& \textbf{0.831} $\pm0.001$ &\textbf{0.850}$\pm0.002 $&\textbf{0.908} $\pm0.005 $\\

\bottomrule
\end{tabular}
}
\caption{Performance of NAO\_ResNet in comparison with other competitors on the BSD500 test set, measured using the method of \cite{xie2015holistically}. For our experiments, the results are presented as mean $\pm std-dev$ computed from 3 runs.}\label{performance_grad}
\end{table}
\newpage
In Table \ref{performance_grad}, our results for \textbf{NAO-Multi-scale} outperform the state-of-the-art results when we use the gradient search space. This is because the gradient can be used to detect edges; hence, using this layer helps the DNN to estimate the edge detection.

\subsection{Semantic Segmentation Task}

\Newcorrection{We propose adapting Deeplab v3+ \cite{chen2018encoder} and learning one deconvolution cell. The Deeplab v3+ backbone is detailed in section \ref{archi_search}.
We train our DNNs on Cityscape \cite{cityscapes}, which comprises 2975 training and 500 validation images. Our results in the validation split are given in Table \ref{performance_seg_city}. We note that the morphological operations help to improve the mIoU on the validation, leading to more accurate DNNs. Moreover, our Deeplab v3+ backbone improves the performance of the classical Deeplab v3+.}
\begin{table}[htbp]
\centering
 \scalebox{0.90}
 {
\begin{tabular}{llc}
\hline
                           & \multicolumn{1}{l|}{}                            & MIou(Cityscapes)                    \\ \hline
 Deeplabv3plus  \cite{chen2018encoder}           & \multicolumn{1}{l|}{Official result}             & 76.9                               \\
NAO-Deeplabv3plus with morpho    & \multicolumn{1}{l|}{arc\_300\_100 batch size=12} & {\color[HTML]{000000} 77.23} \\
NAO-Deeplabv3plus without morpho & \multicolumn{1}{l|}{arc\_300\_100 batch size=12} & {\color[HTML]{333333} 76.97}     \\ \hline
\end{tabular}
}
\caption{\Newcorrection{Performance of NAO-Deeplab v3+ in comparison with the official Deeplab v3+ on the Cityscape \cite{cityscapes} validation set.}}\label{performance_seg_city}
\end{table}

\subsection{Discussion}

We demonstrated in prior experiments that morphological layers boost the DNN's performance for a specific architecture search methodology. However, the improvement is not consistent across all levels. As seen in Table \ref{performance_on_cifar10}, erosion layers perform worse than DNN without morphological layers. However, dilation layers increase performance. This improvement is because erosion is associated with a min-pooling process that extracts less noticeable regions. Similarly, as seen in  Table  \ref{performance_grad}, the DNN learned with dilation layers degrades performance, whereas the gradient-based DNN improves it. This highlights how the correct choice of morphological layers can enhance the representational power of a DNN.
Another intriguing element of the edge detection studies is that the search space of the architecture matters, as we observe that the U-Net search space produces worse results than the multi-scale decoder search space proposed here. As illustrated in Figures \ref{BSD500_FN0} and \ref{BSD500_FP}, multi-resolution aids in obtaining a better edge since some items may be more easily spotted at certain resolutions than others. Moreover, we note that the results obtained using the multi-scale decoder search space with the gradient layer are state-of-the-art, demonstrating the usefulness of these layers. \Correction{Finally, we remark that our strategy of comparing the best architecture with and without morphological layers using NAS takes six days in total, irrespective of which search space we use. An alternative to NAS would be to design an experimental protocol to test all possible architecture configurations. However, such a search would be long and would depend on the exploration protocol.  Hence, our protocol can be used to confirm the interest in using morphological layers.
}

\section{Conclusion}

This paper introduces a new layer for Deep Neural Networks that is based on mathematical morphology. We offer a novel equitable technique for determining the utility of a new layer and apply this method to our newly created layers. We conclude that our layer has the potential to be extremely useful for image categorization and edge detection.
This evaluation methodology appears to be more equitable than the standard approach, which entails proposing a new layer and handcrafting an architecture to increase performance. In this case, everything is optimized using an algorithm.

Additionally, we suggest a new backbone architecture for the architecture search, wherein both the input and output are images; we refer to this as \textbf{NAO-Multi-scale}. Comparing it to the U-net architecture search, we establish that our search architecture shows superior performance.
Finally, we achieve state-of-the-art edge detection performance using \textbf{NAO-Multi-scale} and gradient operations.

In future research, based on these encouraging results, we will examine how to employ such layers to conduct semantic segmentation. Additionally, it will be fascinating to deal with more specialized information, such as remote sensing data containing several photographs of buildings and roads with unique geometric shapes. As such, it may be instructive to observe how these new layers act in this scenario. \Newcorrection{We could also add transformer layers to the search, which was beyond the scope of this paper. We note that these layers show good results in computer vision tasks and aim to study them in future works.}

\noindent\textbf{Acknowledgments:}
\Correction{This work was performed using HPC resources
from GENCI-IDRIS  (Grant 2020-AD011011970) and (Grant 2021-AD011011970R1).}

\bibliographystyle{ieeetr}
\bibliography{mybibfile}

\end{document}